%% file: main.tex
\newif\ifOneCol

\ifOneCol
    \documentclass[12pt,onecolumn,draftcls]{IEEEtran}
    \renewcommand{\baselinestretch}{1.7}
\usepackage{geometry}
\else
    \documentclass[10pt,final,twocolumn]{IEEEtran}
    \renewcommand{\baselinestretch}{1.0}
\fi

\date{}
\usepackage{graphicx}
\usepackage{epsfig}
\usepackage{subfigure}
\usepackage{amssymb}
\usepackage{amsbsy}
\usepackage{amsmath}
\usepackage{steinmetz}
\usepackage[colorlinks=true, pdfborder={0 0 0}]{hyperref}
\usepackage{stmaryrd}
\usepackage{MnSymbol,  bbm} 
\usepackage{cite}
\usepackage{url}
\usepackage{svg}
\usepackage{amsfonts}
   
\usepackage{caption}
\usepackage{color}
\usepackage{float}
\usepackage{times,amsmath,epsfig}
\usepackage{xspace,latexsym,syntonly}
\usepackage{amssymb}
\usepackage{textcomp}
\usepackage{xcolor}
\usepackage{import}

\newtheorem{theorem}{Theorem}

\newcommand{\qed}{\hfill\blacksquare}
\DeclareMathOperator*{\argmax}{\arg\max}

\usepackage{algorithm}
\usepackage{algpseudocode}

\usepackage{pgfplots, pgfplotstable}
\usepackage{tabularray}
\UseTblrLibrary{booktabs}

\usepackage{standalone}

\usepackage{tikz}
\pgfplotsset{compat=1.7}
\usepackage[utf8]{inputenc}

\begin{document}
\title{Multi-Flow Transmission in Wireless Interference Networks:  A Convergent Graph Learning Approach}
\author{Raz Paul, Kobi Cohen \emph{(Senior Member, IEEE)}, Gil Kedar
\thanks{Raz Paul and Kobi Cohen are with the School of Electrical and Computer Engineering, Ben-Gurion University of the Negev, Beer Sheva 8410501 Israel. Email: razpa@post.bgu.ac.il, yakovsec@bgu.ac.il.}
\thanks{Gil Kedar is with Ceragon Networks Ltd., Tel Aviv, Israel. Email: gilke@ceragon.com.}
\thanks{This work was supported by the Israel Ministry of Economy under the Magnet consortium program.}
}

\maketitle

\begin{abstract}
We consider the problem of multi-flow transmission in wireless networks, where data signals from different flows can interfere with each other due to mutual interference between links along their routes, resulting in reduced link capacities. The objective is to develop a multi-flow transmission strategy that routes flows across the wireless interference network to maximize the network utility. However, obtaining an optimal solution is computationally expensive due to the large state and action spaces involved. To tackle this challenge, we introduce a novel algorithm called Dual-stage Interference-Aware Multi-flow Optimization of Network Data-signals (DIAMOND). The design of DIAMOND allows for a hybrid centralized-distributed implementation, which is a characteristic of 5G and beyond technologies with centralized unit deployments. A centralized stage computes the multi-flow transmission strategy using a novel design of graph neural network (GNN) reinforcement learning (RL) routing agent. Then, a distributed stage improves the performance based on a novel design of distributed learning updates. We provide a theoretical analysis of DIAMOND and prove that it converges to the optimal multi-flow transmission strategy as time increases. We also present extensive simulation results over various network topologies (random deployment, NSFNET, GEANT2), demonstrating the superior performance of DIAMOND compared to existing methods.
\end{abstract}


\section{Introduction}
\label{sec:introduction}

The increasing demand for wireless communication services has accompanied the fast development of communication network technology in 5G and beyond. Despite this, spectrum scarcity remains a major constraint in meeting this growing demand. Thus, developing algorithms for data transmission in wireless networks that utilize the available spectral resources and manage data transmissions efficiently is a key challenge in modern wireless communication networks. 

In this paper, we consider the multi-flow transmission problem in wireless interference networks. Specifically, the wireless communication network is modeled by a connected graph, where a communication link between transmitter and receiver is modeled by an edge in the graph. Links cause mutual interference due to the overlapping of their radio frequency signals in the same spatial region. Given $N$ flows, the goal is to allocate multi-flow routes for data transmissions in the wireless network to maximize a certain network utility, where data signals from different flows can interfere with one another due to mutual interference between links along their routes, resulting in a degradation in link capacities.

\subsection{Existing Approaches to Data Flow Transmission}

A well-known approach for data flow transmission is to solve the shortest path problem and use the shortest paths (or a candidate set of short paths) to route data flows. For instance, the popular Open Shortest Path First (OSPF) routing protocol transmits data flows in the network over shortest paths, where the computations of the paths are implemented by the well-known Dijkstra algorithm \cite{srikant2013communication, gong2016distributed}. OSPF-based protocols are widely used in communication companies today due to their simplicity and good performance when the network load is not too high, and the wireless interference is not strong. The drawback of using shortest paths to route data in communication networks is their tendency to increase congestion on those paths, which can decrease performance in highly-loaded networks and in environments with strong wireless interference. Alternatively, more recent approaches have suggested balancing between short paths and path congestion to distribute the load over the network more evenly \cite{ying2010combining, joo2011performance, amar2021online, liu2022routing}. 

In recent years, significant improvements have been made by developing machine learning algorithms for managing flow transmissions in order to tackle uncertainties in random channel and network conditions. Several studies \cite{tekin2011online, tekin2012online, liu2012learning, cohen2014restless, gafni2018learning, bistritz2018distributed, turgay2019exploiting, yemini2020restless, gafni2020learning, gafni2022distributed, gafni2022learning} have analyzed the long-term reward optimization of users in the network using a multi-armed bandit learning framework. The learning strategies have included various methods, such as reinforcement learning and upper confidence bound (UCB)-based algorithms \cite{agrawal1995sample, auer2002finite, tabei2021multi}, as well as deep reinforcement learning that uses deep neural networks in the optimization \cite{wang2018deep, yu2019deep, naparstek2018deep, bokobza2023deep}. While most of these online learning methods have focused on single-hop transmissions, in this paper we apply the learning to multi-hop link states to enable efficient path selection for flow transmission.

Existing learning methods for routing data flows in wireless networks have been developed in several studies \cite{liu2012adaptive, tehrani2013distributed, he2013endhost, talebi2017stochastic, xia2019reinforcement, DRL+GNN, huang2021tsor, Zhao2021, amar2022online}. In \cite{DRL+GNN}, the focus was on using deep reinforcement learning to schedule routes based on congestion levels, but without considering mutual interference between links and without providing theoretical guarantees on convergence. In \cite{liu2012adaptive, tehrani2013distributed, he2013endhost, talebi2017stochastic, xia2019reinforcement, huang2021tsor, Zhao2021, amar2022online}, the focus was on developing an online learning algorithm that efficiently trades off between exploring paths to learn the network state, at the cost of selecting sub-optimal paths during the exploration phase, and exploiting the information the algorithm has gained to solve the shortest paths and schedule data transmissions through these paths. However, these algorithms suffer from poor performance in terms of load balancing since they tend to increase congestion over short paths. This negative effect becomes particularly pronounced as network load increases and mutual interference between links reduces the capacity over selected paths. Our algorithm overcomes this limitation by learning a multi-flow strategy optimized with respect to the current network load, thereby minimizing mutual interference and dynamically adjusting the path selection to reduce congestion in a distributed manner. 

Other related interference-aware methods have been proposed recently in the literature. The Delay-and Interference-Aware Routing (DIAR) algorithm \cite{Chai2020DIAR} is a centralized routing method, where a centralized unit selects routes for all flows. DIAR aims to select routes with minimal end-to-end delay for multiple concurrent data flows. It employs an improved genetic algorithm (GA) to balance the network load and minimize delay. DIAMOND uses machine learning optimization as well, but has several unique aspects that enable substantial enhancements in network performance. First, in contrast to DIAR, which is a fully centralized algorithm, the design of DIAMOND incorporates two consecutive stages, centralized (GRRL) and distributed (NB3R), that leverage the advantages of both centralized knowledge to predict efficient multi-flow route allocation as well as distributed learning to
refine the allocation and significantly improve performance. Second, the GA method in DIAR optimizes its objective online, starting with potentially suboptimal solutions that are refined in subsequent iterations. By contrast, DIAMOND's GRRL provides a pre-trained GNN that yields a near-optimal solution for the distributed stage. This process of pre-training in DIAMOND accelerates the convergence and reduces the time required to reach satisfactory solutions, as evidenced by the simulation results. Third, in contrast to DIAR, DIAMOND has been analytically proven to converge to the optimal allocation over time. Another centralized method, proposed in \cite{He2019JSRC}, is the Joint Source, Routing, and Channel (JSRC) selection scheme. JSRC searches for path allocation in two steps. First, it allocates paths sequentially for each flow, aiming to maximize the capacity along the chosen routes. Next, it refines the paths to further reduce wireless interference by relaxing the bottleneck along a path and replacing it with an alternative route. This approach bears resemblance to a simplified version of our utility function. However, JSRC uses a deterministic search allocation, and does not use a learning-based optimization to efficiently explore the search space as in DIAMOND. Also, it does not incorporate long-term planning for a large number of flows and does not provide convergence guarantee to the optimal allocation, unlike DIAMOND. Among the centralized interference-aware algorithms, DIAR outperformed JSRC in our simulations. As a result, we have presented the results of DIAMOND compared to DIAR, rather than JSRC. A related distributed method is the Interference Aware Cooperative Routing (IACR) \cite{Waqas2022IACR} algorithm. IACR was introduced recently as a distributed routing method for 5G networks. It selects routes for flow demands based on a weighted sum of created and received interference in the network. IACR and DIAMOND share similarities in their objective of reducing interference between routes. However, there are notable differences between the two approaches. Firstly, IACR does not provide performance guarantees as proven by DIAMOND. Additionally, while IACR relies on a simple, non-learnable interference-aware function for optimization, DIAMOND utilizes a GNN as a high-dimensional function approximation \cite{liang2016deep} capable of capturing more complex dynamics between flows. The simulations clearly demonstrate that DIAMOND significantly outperforms IACR. 

In \cite{jiang2020dgn, kortvelesy2022qgnn, liu2020pic}, graph-based approaches using reinforcement learning (RL) have been proposed. Specifically, \cite{jiang2020dgn} introduced a graph convolutional RL method known as DGN, which addresses a multi-agent RL (MARL) problem where agents are represented as nodes in a graph. The objective of DGN is to learn the underlying relations between agents to maximize their average expected return in a cooperative manner. While both DGN and DIAMOND are graph-based RL methods, there are several fundamental differences between the two approaches. Firstly, when applied to our multi-flow transmission problem, DGN treats each flow as an agent and tackles the problem as a multi-agent problem. In contrast, DIAMOND initially approaches the problem from a single-agent RL perspective, and then transitions to a distributed multi-agent perspective in its second stage using the NB3R module. It is important to note that considering each flow as an agent in DGN may lead to longer training times compared to DIAMOND. This is because agents in DGN need to explore the environment to find valid solutions, such as valid routes between source and destination for each flow, before optimizing them. In complex scenarios with a large number of nodes and links, convergence in DGN can be challenging, as commonly observed in MARL. DIAMOND overcomes this issue through its hybrid design of single-agent RL (GRRL) and distributed multi-agent learning (NB3R). Secondly, DIAMOND employs a customized GNN architecture specifically designed to encode paths using its Path Encoder module and capture relations along each link between them. In contrast, DGN captures more abstract relations between flows without explicitly considering their potential transmission routes or the interactions/interferences between links along the routes, focusing more on the overall relations between flows.
Lastly, while DGN is instantiated based on the DQN algorithm, DIAMOND utilizes the REINFORCE algorithm.

Another cooperative MARL approach was proposed in \cite{kortvelesy2022qgnn}, addressing the credit assignment problem in multi-agent frameworks by factorizing the global reward into distributed local rewards, which helps stabilizing the training process. The authors introduced QGNN, a method designed to handle the value function factorization problem. Similar to \cite{jiang2020dgn}, they considered each agent (or flow in our case) as a node in a graph, with edges representing communication channels between agents. Agents receive data from their neighbors to update their policies, utilizing a GRU-based encoding module, similar to our path-encoder. Furthermore, their model incorporates the embedding of agent trajectories to evaluate the Q-value of their actions, resembling our GRRL's embedding of flow paths. It is important to note that \cite{kortvelesy2022qgnn} does not explicitly consider interference over the agent's communication links. The main objective of \cite{kortvelesy2022qgnn} differs from ours: While we aim to solve the multi-flow transmission problem, \cite{kortvelesy2022qgnn} focuses on a MARL problem used to improve the local-global reward factorization between agents. The non-stationarity problem of the MARL environment was studied from a single agent perspective by utilizing a critic that depends on all observations and actions. In settings with multiple agents, this can lead to a challenge where there are $N!$ different possible outputs for the same state due to the permutation of agent orders for the critic. In \cite{liu2020pic}, the authors proposed a permutation invariant critic (PIC) that tackles this issue by being fully centralized, capturing information from all other agents. This centralized deployment of PIC shares similarities with our centralized GRRL module, where global information about all flows is maintained. The difference is that PIC trains all agents in a centralized critic. Thus, although improving scalability compared to previous MARL methods, handling large networks still remains challenging. Finally, \cite{jiang2020dgn, kortvelesy2022qgnn, liu2020pic} considered MARL, each addressing different challenges within the MARL setting. These approaches model agents as nodes in a graph, with edges representing communication relations between them. In contrast, DIAMOND utilizes a GNN to embed the graph structure and information as a communication network signal. Furthermore, \cite{jiang2020dgn, kortvelesy2022qgnn, liu2020pic} do not specifically address wireless interference between links, and some of them are not designed explicitly for communication-related tasks. As a result, extending the centralized GRRL approach of DIAMOND to a fully distributed algorithm could be a potential direction for future work. It is important to highlight that DIAMOND is tailored to the specific setting addressed in this paper, and its hybrid design enables the exploitation of both global and local information. This allows for rigorous convergence guarantees to the optimal solution. In contrast, \cite{jiang2020dgn, kortvelesy2022qgnn, liu2020pic} do not provide theoretical results and convergence guarantees.

\subsection{Main Results}
\label{sec:main_resutls}

In this paper, we aim to solve the multi-flow transmission problem in  wireless interference networks, while overcoming the limitations of existing approaches as discussed earlier. Below, we summarize our main contributions.

Firstly, in terms of algorithm development, our aim is to find a multi-flow transmission strategy that efficiently routes data flows across wireless interference networks while maximizing the network utility. Existing approaches assumed deterministic (e.g., \cite{ying2010combining, srikant2013communication}) or stochastic random (e.g., \cite{liu2012adaptive, tehrani2013distributed, talebi2017stochastic, huang2021tsor, amar2022online}) link weights that allow for efficient learning with low-complexity computation of short paths. However, this assumption is not valid in wireless interference networks, as data signals from different flows can interfere with one another due to mutual interference between links along their routes, leading to a decrease in link capacities. Therefore, finding an optimal solution for this problem is generally computationally expensive due to the large state and action spaces involved in wireless interference networks. 

\begin{figure}[t]
\centering
\includegraphics[scale=0.15]{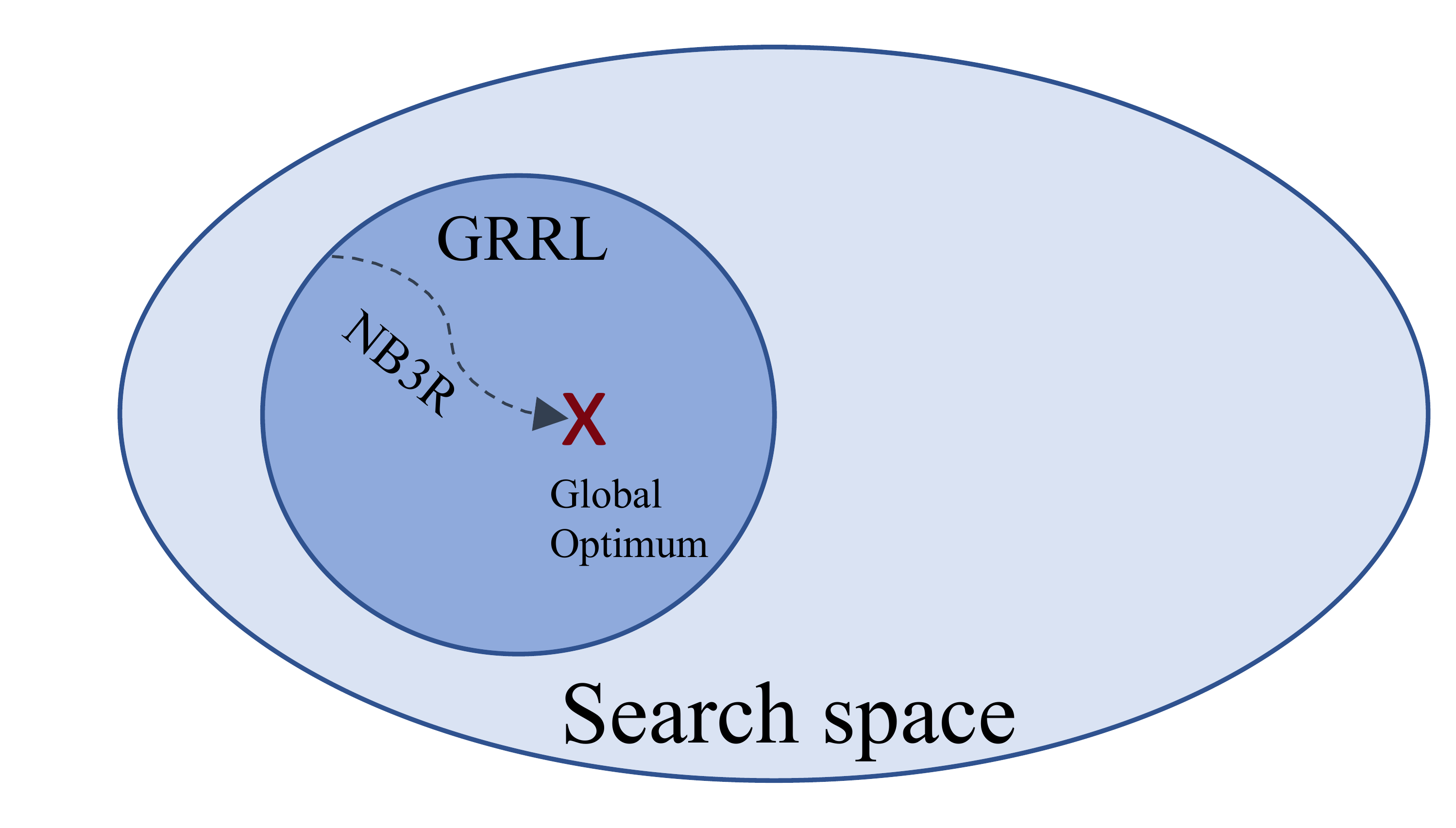}
  \caption{An illustration of DIAMOND's two-stage optimization logic in the search space. The first module, GRRL, applies an efficient search within the large search space and yields a solution that concentrates around the global optimum. The second module, NB3R, avoids local maxima and updates strategies dynamically in a distributed manner to refine the GRRL solution toward a global optimum.\vspace{0.5cm}}
  \label{fig:search_space}
\end{figure}

To tackle this problem, we develop a novel algorithm, dubbed Dual-stage Interference-Aware Multi-flow Optimization of Network Data-signals (DIAMOND). The design of DIAMOND allows for a hybrid centralized-distributed implementation, which is a characteristic of 5G and beyond technologies with centralized unit deployments. The centralized stage computes the multi-flow transmission strategy using a novel module design of Graph neural network Routing agent via Reinforcement Learning, dubbed GRRL. Specifically, the GRRL module is implemented on a centralized unit as OSPF, given the interference map and flow demands, as given in 5G networks. It optimizes the network utility and yields the corresponding path allocation vector, a single route per flow. GRRL is trained by Reinforcement Learning (RL), and uses a novel GNN architecture as its policy network, for efficient path decoding. This captures the ability of GNN to efficiently search the large search space and approximate well the optimal solution. It is designed in a generic way that handles general parameter values, number of nodes, edges, and flows. Then, to avoid local maxima and update strategies dynamically, a distributed stage is implemented to refine the GRRL solution towards a global optimum. To do this, we develop a novel module design of distributed Noisy Best-Response for Route Refinement, dubbed NB3R. In NB3R, each source node updates its path asynchronously in a probabilistic manner to improve a collaborative utility, designed to increase the global network utility. Fig. \ref{fig:search_space} provides an illustration of DIAMOND's two-stage optimization logic in the search space.

Second, we provide a rigorous performance analysis, theoretically and numerically. In terms of theoretical performance analysis, we provide a rigorous convergence analysis of the DIAMOND algorithm. Specifically, we prove that DIAMOND converges to the optimal multi-flow path allocation that maximizes the global network utility as time increases. In terms of numerical performance analysis, we present extensive simulation results to evaluate the overall performance of DIAMOND in finite time. The simulations were conducted using various network topologies, namely random deployment, NSFNET, and GEANT2 networks. All simulations validated the significant performance improvements of DIAMOND compared to existing methods.

\begin{figure}[ht]
\centering
\includegraphics[width=0.7\linewidth]{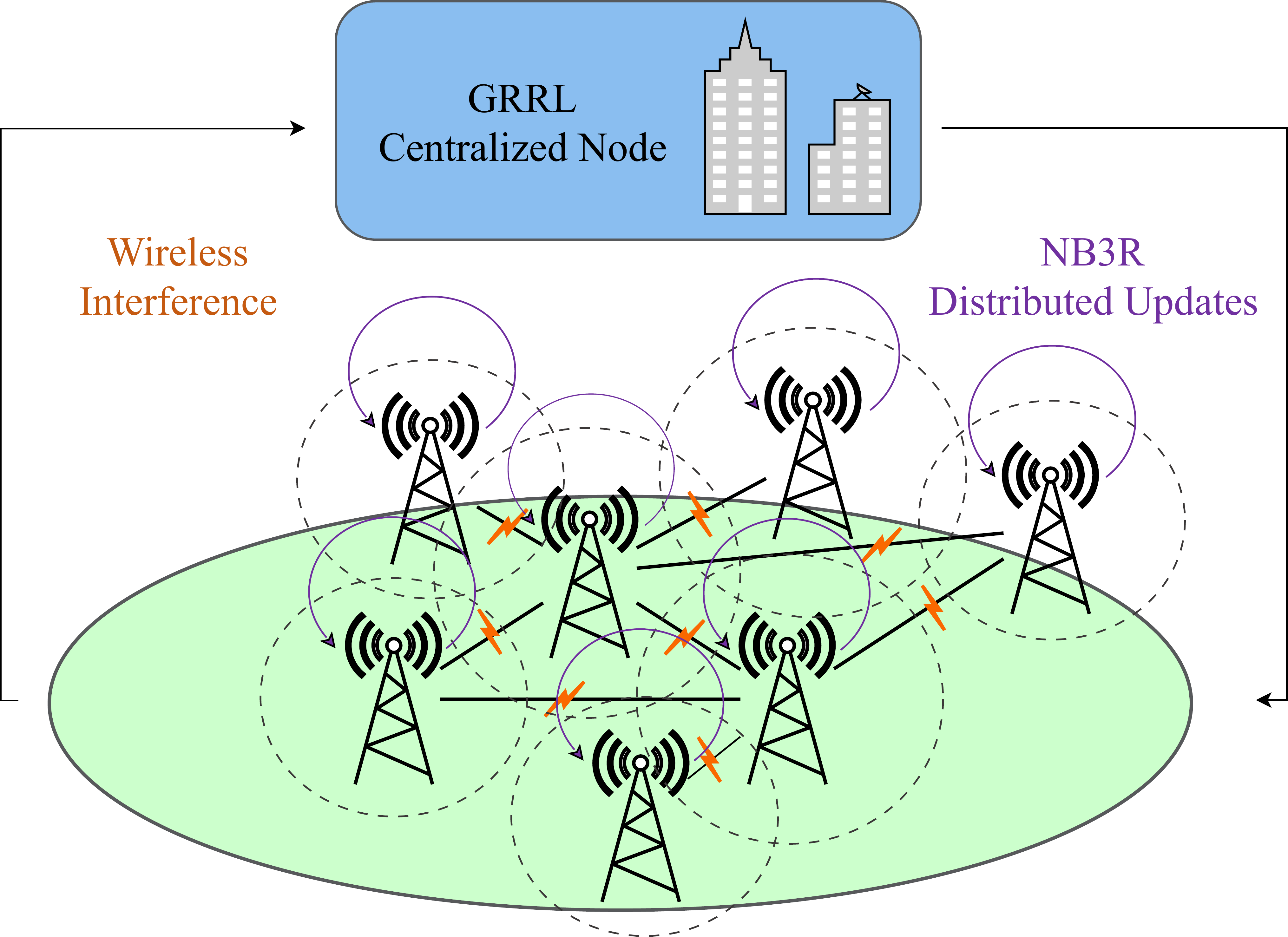}
  \caption{An illustration of the system model that allows for a hybrid centralized-distributed implementation via DIAMOND. The wireless interference between links in the network is visually illustrated. A centralized node runs GRRL to generate an initial path allocation using global information about the network's state. Then, the nodes refine their allocation in a distributed manner using NB3R algorithm.}
  \label{fig:system_overview}
\end{figure}

\section{Network Model and Problem Statement}
\label{sec:system}

The system model allows for a hybrid centralized-distributed implementation via DIAMOND. In this section we explain the system model in detail. An illustration is presented in Fig. \ref{fig:system_overview}. We consider the multi-flow transmission problem in wireless interference networks as described next. Let $\mathcal{V}$ denote the set of nodes (i.e., users) in the network, and $\mathcal{E}$ the set of edges (i.e., communication links between nodes). The wireless communication network is modeled by a directed connected graph $G=(\mathcal{V},\mathcal{E})$. A directed communication link between transmitter $v$ and receiver $v'$ is modeled by a link $(v,v')\in \mathcal{E}$ from node $v\in\mathcal{V}$ to node $v'\in\mathcal{V}$ in the directed graph (i.e., $v,v'$ are neighbors). Recall that links cause mutual interference due to the overlapping of their radio frequency signals in the same spatial region. Let $\mathcal{F}$ be a set of $N$ data flows, where each flow (say $f_n\in\mathcal{F}$) is indicated by source node $s_n\in\mathcal{V}$ and destination node $d_n\in\mathcal{V}$. Let $\mathcal{A}_n=\{\varphi^{n}_1,\varphi^{n}_2,\ldots,\varphi^{n}_{K_n}\}$ be the set of all allowed routes for flow $n$. Let $\sigma_n\in\mathcal{A}_n$
be a selected route for flow $n$, $\sigma\triangleq(\sigma_1, \sigma_2, ..., \sigma_N)$ be the selected route vector for all flows, and $\sigma_{-n}\triangleq(\sigma_1, ..., \sigma_{n-1}, \sigma_{n+1},..., \sigma_N)$ be the selected route vector for all flows excluding flow $n$. Let $u_n(\sigma)$ be a bounded utility of flow $n$ (e.g., the achievable rate, a monotonically increasing function with the achievable rate, etc.), $u_n(\sigma)\leq u_{max}\;\forall n=1, 2, ..., N$. Note that the utility depends on the selected route of flow $n$ as well as the selected routes of other flows that interfere with flow $n$. We denote the set of flows that interfere with flow $n$ (i.e., interfering neighbors) by $\mathcal{N}(n)$. All other flows $\mathcal{F}\setminus\mathcal{N}(n)$ are transmitted through paths that cause interference below the noise floor (due to geographical distance, directional antennas, etc.). As a result, $u_n(\sigma)$ depends on the selected route of flow $n$, and the selected routes of flows $\mathcal{N}(n)$, $u_n(\sigma)=u_n(\sigma_n, \left\{\sigma_i\right\}_{i\in\mathcal{N}(n)})$.

The objective is to find a multi-flow 
transmission path vector $\sigma$ that solves the network utility maximization (NUM) problem 
\cite{srikant2013communication}:  
\begin{equation}
\label{eq:objective}
    \sigma^* = \argmax_{\{\sigma_n\in\mathcal{A}_n\}_{n=1}^N} \sum_{n=1}^N u_{n}(\sigma). 
\end{equation}
The utility of flow $n$ is typically set to be a monotonically increasing function with the achievable rate $R_n(\sigma)$ \cite{srikant2013communication}. For instance, one can set $u_n(\sigma)=R_n(\sigma)$ to maximize the network sum rate, or $u_n(\sigma)=\log R_n(\sigma)$ to maximize the network sum log-rate (i.e., proportional fairness). The rate of flow $f_n$ transmitted through path $\sigma_n=(s_n, v_1, v_2, ..., d_n)$ is determined by the slowest link rate across the links in its path, say link $\ell$. The achievable rate of user $n$ is thus given by:
\begin{equation}
    R_n(\sigma)=R_{\ell}(\sigma)=
    B_\ell \cdot \log\left(1+\textrm{SINR}_\ell(\sigma)\right), 
    \label{eq:link_rate}
\end{equation}
where $B_\ell$ is the bandwidth of link $\ell$, and 
\begin{equation}
 \textrm{SINR}_\ell(\sigma)\triangleq\frac{P_\ell}{I_\ell(\sigma)
 +\tilde{\sigma}^2}   
\end{equation}
denotes the Signal to Interference plus Noise Ratio (SINR) at the reciever of link $\ell$. Here, $P_\ell$ is the received power of the signal, $\tilde{\sigma}^2$ is the power spectrum density (PSD) of the additive white Gaussian noise (AWGN), and 
\begin{equation}
I_\ell(\sigma) \triangleq \sum_{i\in\mathcal{N}_{\ell}}
I_{\ell,i}(\sigma_i)
\end{equation}
is the cumulative interference power at the receiver of link $\ell$, aggregated across all links within its interference range $\mathcal{N}_\ell$. Here, $I_{\ell,i}$ is the interference power at the receiver of link $\ell$ caused by flow $i$, greater than zero if active by strategy $\sigma_i$. The transmission power control mechanism considered in this paper is independent, where a predetermined transmission power is assigned to each link based on factors such as geographical distance and channel state. Consequently, the action space of the learning algorithm is solely determined by the selected paths that affect the network utility.

Finding the optimal multi-flow transmission path vector that solves the combinatorial optimization in (\ref{eq:objective}) is an NP-hard problem \cite{NP_hard_routing}. The exponential number of possible route allocations makes it intractable to find the optimal solution via naive brute-force methods, thus a computationally-efficient algorithm is needed. Moreover, the optimal allocation is often expensive to compute, even for a small network. Thus, developing learning-based methods by training a supervised model is not practical. In the next section, we develop a novel algorithm to solve (\ref{eq:objective}), while overcoming these limitations. 

\begin{figure*}
    \vspace{-1cm}
    \centering
    \includegraphics[scale=0.15]{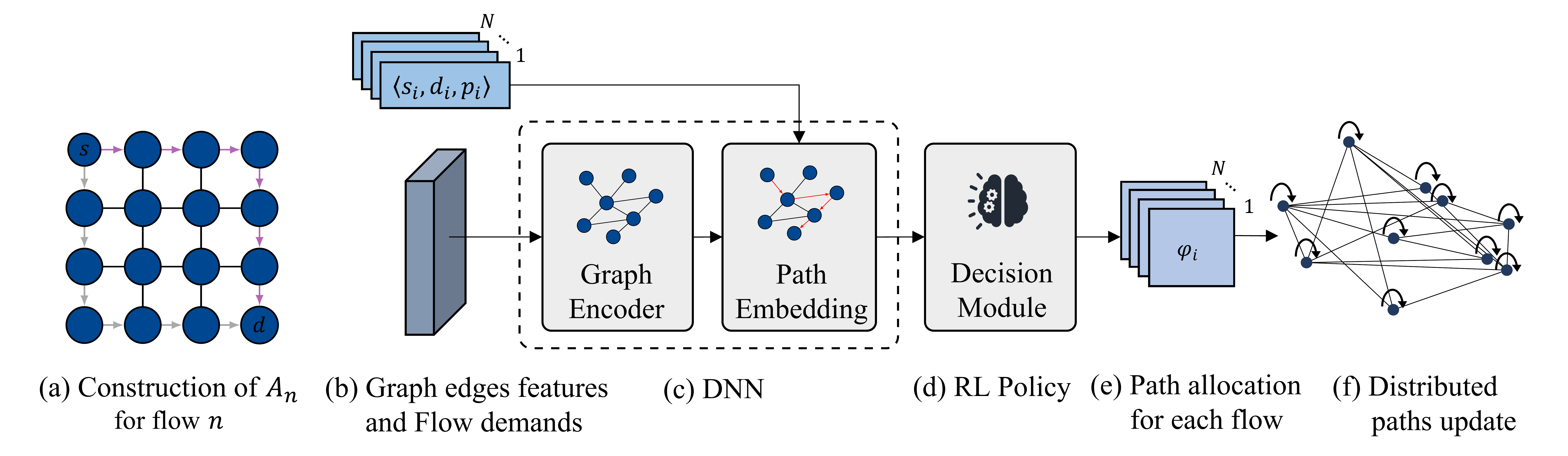}
    \caption{An overview of the proposed DIAMOND framework: (a) An illustration of the construction of the search space for flow $n$, as described in Subsection \ref{ssec:constructing} and Alg.\ref{alg:k_paths}. This stage is implemented for all flows $n=1, 2, ..., N$. (b)-(e) \emph{Centralized GRRL module:} (b) The RL agent receives $N$ flow demands, and the network state as link features. (c) The network state is processed with a graph encoder GNN to produce an embedding for each link as well as a global graph embedding. The flow demands, along side those embedding are processed by the path-embedding module that outputs an embedding for each of the possible routing options. (d) The RL agent makes an allocation decision based on the path embedding. (e) The transmission paths of all $N$ flows are allocated at once to the network. (f) \emph{Distributed NB3R module:} Each flow updates its path allocation distributedly, based on the NB3R policy, which refines the allocation made by GRRL to improve performance.}
    \label{fig:alg_overview}
\end{figure*}

\section{The Proposed DIAMOND Algorithm}
\label{sec:DIAMOND}

In this section we present the DIAMOND algorithm to solve the objective (\ref{eq:objective}). The design of DIAMOND allows for a hybrid centralized-distributed implementation, which is a characteristic of 5G and beyond technologies with centralized unit deployments. 
DIAMOND is a two-stage approach consisting of the GRRL stage (centralized) and NB3R stage (distributed). In the first stage, using GRRL, a single-agent RL method is used to produce the allocation for all $N$ flows simultaneously. In the second stage, using NB3R, a distributed multi-agent learning is used to refine the allocation. In this stage, at each time step, only the route for one single flow is selected, which updates its route allocation.

The centralized stage computes the multi-flow transmission strategy using deep reinforcement learning methodology that uses deep neural network (DNN) to capture the high-dimensional problem. Specifically, we design a novel module of GNN Routing agent via Reinforcement Learning, dubbed GRRL. The GRRL module is implemented on a centralized unit as OSPF, given the interference map and flow demands, as given in 5G networks. It optimizes the network utility and yields an initial path allocation vector with fast inference computation. GRRL is trained by RL offline, and uses a novel GNN architecture as its policy network, for efficient path decoding. This captures the ability of GNN to efficiently search the large search space and approximate well the optimal solution. It is designed in a generic way that handles general parameter values (number of nodes, edges, and flows). Then, to avoid local maxima and update strategies dynamically, a distributed stage is implemented to refine the GRRL solution toward a global optimum. To do this, we develop a novel module design of distributed Noisy Best-Response for Route Refinement, dubbed NB3R. In NB3R, each source node updates its path asynchronously in a probabilistic manner to approach the global NUM solution \eqref{eq:objective}. A high-level illustration of DIAMOND's two-stage optimization logic in the search space is presented in Fig. \ref{fig:search_space}. A detailed illustration of each module in DIAMOND's framework is presented in Fig. \ref{fig:alg_overview}. 

We will begin by discussing the construction of the search space in the DIAMOND algorithm, which enables efficient learning. Following that, we will provide a detailed explanation of the GRRL and NB3R modules utilized in DIAMOND.

\subsection{Constructing the Search Space in the DIAMOND Algorithm}
\label{ssec:constructing}

The number of possible routes for each flow is usually large, resulting in a large action space for multi-flow path allocation that increases exponentially with the number of flows $N$. Therefore, to prevent divergence of the learning algorithm, the search space (i.e., the allowed number of paths for each flow) must be restricted \cite{babaee2010cross}. Let $K$ be the number of allowed paths for each flow. Then, the size of the action space  is $K^N$ that determines the search space of the algorithm. Generally, one can set $K_n$ as the number of allowed paths for flow $n$, which can vary across flows. In \cite{DRL+GNN}, the authors used the standard selection of $K$ shortest-paths to restrict the search space. However, this approach is not suitable for wireless interference networks, as the selection mechanism may choose $K$ short paths with high mutual interference. Therefore, to cope with the interference environment, we develop a novel heuristic search space reduction method, which is described next (the pseudocode is given in Algorithm \ref{alg:k_paths}). For each flow, the algorithm computes a set of $K$ short paths that have high mutual distance between them. This allows the algorithm to select paths with significantly different interference effects on the network to effectively balance the load across the network. Specifically, let $d(\ell, \varepsilon)$ be a certain distance measure between links $\ell$ and $\varepsilon$ (e.g., Manhattan distance, Euclidean distance). The links' weights are initialized by a pre-defined function denoted by $c(\varepsilon)$, that can be set based on various factors such as the SINR on that link, the channel gains, or user preferences for prioritized links. The first shortest path (say $\varphi$) is computed by the Dijkstra algorithm (the function shortest-path$(s_n, d_n, G, \left\{W(\varepsilon)\right\}_{\varepsilon\in\mathcal{E}})$ in Algorithm \ref{alg:k_paths}). Then, the algorithm increases the weight of each link (say $\varepsilon$) by $\min_{\ell \in \varphi} d^{-1}(\ell, \varepsilon)$. As a result, the next selected shortest path by Dijkstra (say $\varphi'$) would have a large distance between $\varphi'$ and $\varphi$. The algorithm continues until $K_n$ paths are selected.

\begin{algorithm}[hbt!]
\caption{Construction of $\mathcal{A}_n$ for flow $n$}\label{alg:k_paths}
\begin{algorithmic}[1]
\State \textbf{Input:} $G=(\mathcal{V}, \mathcal{E}), \ s_n, d_n \in \mathcal{V}$ 
\State \textbf{Initialize:} $W(\varepsilon) \gets c(\varepsilon)\;  \forall \varepsilon \in \mathcal{E}$, \; $P \gets [] $
\For{$i=1,2,\ldots,K_n$ }
\State $\varphi \gets \textrm{shortest-path}(s_n, d_n, G, \left\{W(\varepsilon)\right\}_{\varepsilon\in\mathcal{E}})$
\State $P \gets P \cup \{\varphi\}$
\State  $W(\varepsilon) \gets W(\varepsilon) + \min_{\ell \in \varphi} d^{-1}(\ell, \varepsilon), \forall \varepsilon \in \mathcal{E}$
\EndFor
\State \textbf{Return:} $P$
\State $\mathcal{A}_n\gets P$
\end{algorithmic}
\end{algorithm}

\subsection{The Centralized Stage (GRRL)}

We start by describing the DNN architecture used in the GRRL module. Then, we describe the DRL algorithm of the agent. 

\subsubsection{The DNN architecture} \label{sec:DNN_arch}

The agent's policy is implemented as a deep GNN, which is comprised of graph-encoder and path-encoder modules.

\noindent
\textbf{Graph-Encoder}: The first step involves implementing Message Passing Neural Networks (MPNN) with a customized message-passing procedure that considers both the uplinks and downlinks across the links in the network. The input is an $E \times h_{\textrm{in}}$ edge feature matrix, containing all feature vectors of the graph's edges, and the $N$ flow demands triplets. Here, $h_{\textrm{in}}=3$, consists of the interference on the link, the link's capacity (normalized), and the last action that was taken (one-hot representation). We treat the links in the communication graph as nodes in the GNN. Let $\ell = (v,v')\in\mathcal{E}$ be a link in the communication network. We define $E_{in}(\ell) = \{(w,v) \ | \ w\in\mathcal{V} ,(w,v)\in\mathcal{E}\}$, and $E_{out}(\ell) = \{(v',w) \ | \ w\in\mathcal{V} ,(v',w)\in\mathcal{E}\}$
to be the sets of \emph{incoming} and \emph{outgoing} edges for link $\ell$.
The term $h(\ell)$ denotes the embedding for link $\ell$, and superscripts denote the time-step indices of the message passing. A single message passing can be described as 
\begin{eqnarray}
    h^{(t)}(\ell) &=& \gamma \big(h^{(t-1)}(\ell)W_1^{t}+ \sum_{e\in E_{in}(\ell)}h^{(t-1)}(e)W_2^{(t)}  \nonumber \\ 
    &+& \sum_{e\in E_{out}(\ell)}h^{(t-1)}(e)W_2^{(t)} \big), \label{eq:custom-message-passing}
\end{eqnarray}
where $W_1, \ W_2, \ W_3$ are learned matrices, and $\gamma(\cdot)$ denotes a non-linear activation function. As implemented by MPNNs, \eqref{eq:custom-message-passing} is held for $T$ iterations. In our case, the \emph{update} step is implemented as a simple GRU \cite{GRU}. This module results with an embedding representation $h(\ell)\in\mathbb{R}^{d}, \ \forall \ell\in\mathcal{E}$, as well as a global graph embedding $h(G)\in\mathbb{R}^{d}$, as the mean embedding.
Specifically, the term $h(G)$ represents the global graph embedding, which is obtained by applying the Graph-Encoder module to each individual link in the graph. For each link $\ell \in \mathcal{E}$, the Graph-Encoder module outputs an embedding vector $h(\ell) \in \mathbb{R}^d$, where $d$ is the dimensional of the embedding. These individual link-embeddings are then aggregated to obtain the global graph embedding $h(G)$. The aggregation operation used to compute $h(G)$ is the  permutation-invariant mean operation, which calculates the average embedding vector across all the links in the graph. Mathematically, $h(G)$ is computed as follows: $h(G)=\frac{1}{E}\sum_{\ell \in\mathcal{E}} h(\ell) \in \mathbb{R}^{d}$, where $E$ is the total number of links in the graph.

\noindent
\textbf{Path-Encoder}: Next, an embedding $\eta_i$ per each path $\varphi_i^n$ in the action space $\mathcal{A}_n$ is constructed. This module is implemented as a bi-directional GRU. Its hidden state is initialized by $h(G)$. For each path in the action space, it sequentially aggregates all the edge features along the path with the flow demand embedding, and outputs a path specific embedding $\eta_i, \ i=1,2,\ldots,K_n$.

Finally, the agent determines the path to be allocated, by sampling the probability computed by scaling $\eta_i$ using the softmax operator 
\begin{equation}
    p_i^{n} = \frac{\exp (\eta_i)}{\sum_{j=1}^{K_n} \exp(\eta_j)} \label{eq: softmax}
\end{equation}
based on those embeddings.

As mentioned before, links of the communication graph are treated as nodes in the GNN. We note that existing routing protocols also have complexity that depends on the number of edges. For instance, OSPF, which uses Dijkstra algorithm, has a complexity of $O(E+V\log V)$. While DIAMOND requires a GNN implementation that depends on the network size, recent studies have demonstrated efficient implementations of GNNs \cite{DRL+GNN, Shen2021Scalable, liu2022survey, besta2022parallel, Luo2022reram, Zhang2021fccm, Zheng2022bytegnn}, making them a promising method for solving network optimization problems and applicable in the context of 5G networks.  It is important to highlight that training the GNN is performed offline. We leverage the advantageous property of GNNs, which is their ability to transfer knowledge between different graphs of varying sizes and topologies. This allows the trained GNN to generalize well to unseen network sizes and topologies during execution.

Finally, we point out that if scalability becomes a concern, there is an option to divide the network into smaller clusters. This approach allows for the decoupling and separate optimization of routes within each cluster and between clusters, similar to the existing practice in the internet where IGP protocols handle routing data within autonomous systems (ASs) and EGP protocols handle routing data between ASs. By employing this strategy, the complexity of route optimization can be significantly reduced.

\subsubsection{The Operation of the DRL Agent}

To implement the DRL algorithm for the agent, we design the reward function judiciously so that to balance between the utility of a flow and its interference to other flows. Since we aim at maximizing the sum of flow-utility in the network, we design the reward signal as the difference between two consecutive rewards, i.e., $R_t = r_t - r_{t-1}$, where $r_t$ is the reward given from the environment, and $R_t$ denotes the reward introduced to the agent at time-step $t$. We have achieved a return as a telescopic sum. By setting $r_{0}=0$, we have:
\begin{equation}
    \mathcal{R}(\sigma) = \sum_{n=1}^N R_n = \sum_{n=1}^N (r_{n}-r_{n-1}) = 
    r_{N}. \label{eq:trajectory_return}
\end{equation}
For example, let $a_n$, and $\Phi_n$ denote the action, and the set of all allocated paths at time step $t=n$. Then, a good reward function that balances between the individual utility and the interference level in the network can be defined by: $r_n(a_n)=\alpha\cdot \sum_{m=1}^{n} u_m(\sigma)-(1-\alpha)\cdot\sum_{\ell\in\mathcal{E}(\Phi_n)} I_{\ell}(\sigma)$.
We note that the additional interference term presents a general example. In the case where the utility $u_n$ is set to optimize the proportional fairness (i.e., to maximize sum-log rates) or sum-rates, it is redundant, and thus one can set $\alpha$ to $1$ to eliminate the additional interference term.

We train the agent using REINFORCE with baseline algorithm \cite{REINFORCE}. 
REINFORCE is a policy gradient method, which uses Monte Carlo rollout to compute the rewards, i.e., play through the whole episode to produce an allocation to each flow, then receives the trajectory's reward $\mathcal{R}(\sigma)$, given in \eqref{eq:trajectory_return}, which we treat as the agent's loss $\mathcal{L}$. Given the multi-flow transmission problem, the DNN model parameterized by $\boldsymbol{\theta}$ produces probability distribution $p_{\boldsymbol{\theta}}(\pi)$ by \eqref{eq: softmax}. It is then sampled to obtain a path allocation $\sigma=(\varphi_n\in\mathcal{A}_n)_{n=1}^{N}$. With the actions $\sigma$ given a network state $s$, the REINFORCE gradient estimator minimizes $\mathcal{L}_{\boldsymbol{\theta}}$ using the policy gradient:
\begin{eqnarray}
\nabla_{\boldsymbol{\theta}}\mathcal{L}_{\boldsymbol{\theta}} = \mathbb{E}_{p_{\boldsymbol{\theta}}(\sigma)}[\nabla \log p_{\boldsymbol{\theta}}(\sigma) (\mathcal{R}(\sigma)-b(s))],
\end{eqnarray}
where $b(s)$ is the baseline. In practice, we update $\boldsymbol{\theta}$ by Adam optimizer according to $\nabla\mathcal{L}$. The baseline is set to be the best random choice (one path for each flow) out of $M$ i.i.d. trials, which is only a function of the graph state, and does not depend on the action taken by the agent. Thus, it does not introduce any bias to the gradient estimator \cite{BartoSutton}. The pseudocode code of the operation of the DRL agent is given in Algorithm \ref{alg:RL}. Given a set of $N$ flow demands, the DRL agent allocates all $N$ flow demands one after the other, in a random order, by interacting with the simulated network environment.
First, as an initialization (line 3), it observes the network's initial state $\mathcal{G}$ as the $E \times h_{\textrm{in}}$ edge feature matrix discussed above. For each flow, say $n$, the agent first calculates its action space $\mathcal{A}_n$, namely, the $K_n$ possible paths for allocation (line 5). Then, the agent generates an embedding for each possible path $\pi_n$ (line 6). The embedding is produced via the GNN in a two-stage process, where first the Graph-Encoder generates features for each edge in the graph given the current state $\mathcal{G}$, followed by the Path-decoder module which produces embedding for each possible path in $\mathcal{A}_n$. Hence, $\pi_n$ is a vector of size $K_n$. The agent then chooses the path to allocate for flow $n$ by sampling $\pi_n$ in \eqref{eq: softmax} (line 7). Lastly, it allocates the path to the environment, and receives in return a new state of the environment, following its allocation action (line 8).

\begin{algorithm}
\caption{The operation of the DRL agent}\label{alg:RL}
  \begin{algorithmic}[1]
    \State \textbf{Inputs:} $G=(\mathcal{V}, \mathcal{E}), \{s_n, d_n\}_{n=1}^{N}, \{K_n\}_{n=1}^{N}$
    \State \textbf{Output:} $\{\varphi_n\}_{n=1}^{N}$
    
    \State $\mathcal{G} \gets \textrm{env.reset}()$
    \For{$n=1, 2, \ldots, N$}
    \State $\mathcal{A}_n \gets \textrm{Alg}.\ref{alg:k_paths}(s_n, d_n, K_n)$
    \State $\textbf{$\pi_n$} \gets \textrm{GNN}(\mathcal{A}_n, \mathcal{G})$
    \State $\varphi_n \sim Softmax(\pi_n)$
    \State $\mathcal{G} \gets \textrm{env.allocate}(\varphi_n)$
    \EndFor
    \State \textbf{Return:} $\{\varphi_n\}_{n=1}^{N}$
\end{algorithmic}
  \end{algorithm}

\subsection{The Distributed Stage (NB3R)}
\label{subsec:analyze_stage_2}

After a route was chosen for each flow (as illustrated in Fig.\ref{fig:alg_overview}(e)), an iterative noisy best-response procedure is implemented distributedly to refine the path chosen by the GRRL stage. 

Theoretically, convergence analysis often requires flows to update their strategies sequentially to avoid conflicts that might arise from simultaneous updates. In communication systems, this is often achieved by allowing each source node to draw a random backoff time and update its strategy when the backoff time expires \cite{srikant2013communication, cohen2017distributed}. For simplicity, we assume a similar mechanism here. Specifically, we assume that flows hold a global clock and may update their strategies only at certain times $\tilde{t}_1, \tilde{t}_2, ..., $ referred to as updating times. At each updating time, every flow draws a backoff time from a continuous uniform distribution over the range $[0, \tilde{T}]$ for some $\tilde{T}>0$. A flow whose backoff time expires may broadcast a pilot signal to its neighbors, indicating that its strategy has been updated. Then, all its neighbors keep their strategies fixed until the next updating time. As a result, neighbors will not update their strategies simultaneously, as desired. At each updating time, we refer to flows that update their strategies to improve the overall objective as \emph{optimizing flows}, denoted by $\mathcal{F}{opt}$ (which is time-varying across updating times). Algorithm \ref{alg:stage_2} presents the pseudocode of the NB3R module.

Let us define a collaborative utility for each flow as follows:   
\begin{equation}
    \mathcal{U}_{n}(\sigma) = u_n(\sigma) + \sum_{m\in\mathcal{N}(n)} u_m(\sigma). 
\label{eq:utility}
\end{equation}
Note that the collaborative utility $\mathcal{U}_{n}(\sigma)$ consists of the individual utility $u_{n}(\sigma)$ of flow $n$ plus the utilities of all flows that interfere with flow $n$.  

The NB3R procedure works as follows. The distributed learning stage is initialized by the path allocation computed by the GRRL module. Then, a set of optimizing flows $\mathcal{F}_{opt}$ is selected (line 5). Each optimizing flow (say flow $n$) in $\mathcal{F}_{opt}$ queries its neighbors $\mathcal{N}(n)$ to compute and send their individual utility $u_m(\varphi, \sigma_{-n})$ for $K_n$ possible values, $\sigma_n=\varphi_1, ..., \varphi_{K_n}$ given the current strategy profile of all other flows $\sigma_{-n}$ (line 7). Note that each neighbor only needs to know the strategy of its neighbors to compute the utility values. Then, flow $n$ computes the collaborative utility using (\ref{eq:utility}) (line 8). 

Next, the path is updated using noisy best response (NBR) dynamics \cite{young2020individual}. Unlike in best response (BR) dynamics, where a player updates its strategy to maximize its utility given the current strategy of all other players, in NBR a player updates its strategy in a probabilistic manner. This allows for the avoidance of local maxima when optimizing the overall objective. Here, optimizing flows construct a probability mass function (pmf) over their actions (i.e., possible paths) and choose their actions according to this distribution. Typically, NBR is designed so that the BR strategy is played with high probability, while other strategies are played with a probability that decays exponentially fast with the myopic utility loss in order to escape local maxima. Specifically, the pmf over the available actions is given by:
\begin{equation}
    \mathbb{P}(\varphi^n_i) = \frac{e^{\nu \cdot \mathcal{U}_n(\varphi^n_i, \varphi^{-n})}}{\sum_{j=1}^{|\mathcal{A}_n|} e^{\nu \cdot \mathcal{U}_n(\varphi^n_j, \varphi^{-n})}} 
    \label{eq:nb3r_pmf}
\end{equation}
for some $\nu>0$. 
Note that when $\nu=0$, the pmf assigns equal weights for all strategies, while the probability of playing BR approaches one as $\nu\rightarrow\infty$ (a discussion on the setting of $\nu$ over time based on simulated annealing analysis \cite{hajek1988cooling} is provided in the next section).

After updating the selected path (line 9), flow $n$ broadcasts this information to all its neighbors $m\in\mathcal{N}(n)$ (line 10). The procedure repeats until convergence (convergence analysis is provided in the next section).   

NB3R requires communication only between neighbors during the strategy update, rather than with the entire network. This aspect provides a significant advantage in learning algorithms. when a flow (say $n$) changes its strategy, it can broadcast a pilot signal to its neighbors, indicating that its strategy has been updated. The neighbors can then update their utility based on this information. This approach helps in reducing the communication overhead while still allowing the nodes to make informed decisions based on the latest strategy updates.

\begin{algorithm}[hbt!]
\caption{NB3R Algorithm}\label{alg:stage_2}
\begin{algorithmic}[1]
\State \textbf{Inputs:} $G=(\mathcal{V}, \mathcal{E}), \{s_n, d_n, \mathcal{A}_n\}_{n=1}^{N}$, path allocation $\{\varphi_n\}_{n=1}^{N}$ output by  \textrm{Alg. \ref{alg:RL}} \vspace{0.1cm}
\State \textbf{Output:} Path $\varphi^{*}_n$ for each flow $n$ (distributedly)
\State \textbf{Initialize:} Each flow (say $n$) transmits its data signal through path $\varphi_n$
\Repeat{\;(at each updating time)} 
\State $\mathcal{F}_{opt} \gets \textrm{updated set of optimizing flows}$
\For{$n\in\mathcal{F}_{opt}$}:
\State Query $K_n$ utility values: 
\Statex \hspace{1cm}$u_m(\varphi_1^n, \sigma_{-n}), \ \ldots, \ u_m(\varphi_{K_n}^n, \sigma_{-n})$ 
\Statex \hspace{1cm}from all neighbors $m\in\mathcal{N}(n)$ 
\State Compute $\mathcal{U}_{n}(\varphi_1^n, \sigma_{-n}), \ \ldots, \ \mathcal{U}_{n}(\varphi_{K_n}^n, \sigma_{-n})$ (\ref{eq:utility})
\State \textrm{Draw $\varphi_n$ from distribution \eqref{eq:nb3r_pmf}} 
\State \textrm{Inform all neighbors $m\in\mathcal{N}(n)$ of the updated $\varphi_n$}
\EndFor 
\Until all utilities converge
\State \textbf{Return:} $\{\varphi^{*}_n\}_{n=1}^{N}$
\end{algorithmic}
\end{algorithm}

\section{Performance Analysis}

In this section, we evaluate the performance of the DIAMOND algorithm. First, in Subsection \ref{ssec:theoretical}, we provide theoretical convergence analysis, demonstrating strong guarantees that the algorithm will converge asymptotically to the optimal multi-flow path allocation as time increases. Then, 
in Subsection \ref{ssec:numerical} we present extensive simulation results to evaluate the overall performance of DIAMOND numerically in finite time. The
simulations were conducted using various network topologies,
namely random deployment, NSFNET, and GEANT2 networks.

\subsection{Theretical Analysis}
\label{ssec:theoretical}

In the following theorem we show that DIAMOND converges to the optimal multi-flow path allocation asymptotically as time increases. Let $\sigma^{DIAMOND(\nu)}(t)$ be the strategy profile under DIAMOND at time $t$ given a parameter value $\nu$. Let 
$\Sigma^*$ be the set of strategy profile that solves \eqref{eq:objective}. The following theorem shows that by setting $\nu$ sufficiently large, the probability of reaching the optimal strategy profile is close to one as time increases.\vspace{0.1cm}   

\noindent
\begin{theorem}
\label{th:main_theorem}
For any $\epsilon>0$, there exists $\nu>0$ such that
\begin{equation}
\label{eq:prob_nb3r}
    \lim_{t\to\infty} \mathbb{P}\left(\sigma^{DIAMOND(\nu)}(t)\in\Sigma^*\right) \geq 1-\epsilon.\vspace{0.3cm} 
\end{equation}
\end{theorem}

\emph{Proof:}
We prove the theorem by using potential game analysis \cite{potential_games}. We refer to each flow as a player that can adjust its strategy (i.e., selected path). The game is initialized by the centralized solution (GRRL), and plays distributedly by the NB3R dynamics. In potential games, the incentive of  players to switch strategies can be
expressed by a global potential function. A Nash Equlibrium for the game is reached at any local maximum
of the potential function. Let $\sigma_{n}^{(1)}$, $\sigma_{n}^{(2)}$ be two possible strategies for flow $n$. Specifically, the DIAMOND dynamics follows an \emph{exact} potential game if there is an exact potential function $\phi: \sigma \to \mathbb{R}$ such that for every flow $n$ and for every $\sigma_{-n}=\{{\varphi^{i}}\}_{i\neq n}$, the following holds:
\begin{equation}
\begin{array}{l}
\label{eq:definition_exact}
\displaystyle
\mathcal{U}_n(\sigma_{n}^{(2)}, \sigma_{-n}) -\mathcal{U}_n(\sigma_{n}^{(1)}, \sigma_{-n}) \vspace{0.2cm}\\
\displaystyle\hspace{2.5cm}
=\phi(\sigma_{n}^{(2)}, \sigma_{-n}) - \phi(\sigma_{n}^{(1)}, \sigma_{-n}),
\vspace{0.2cm}\\
\displaystyle\hspace{4cm}
\forall\sigma_{n}^{(1)}, \sigma_{n}^{(2)}\in\mathcal{A}_n.
\end{array}    
\end{equation}

Next, we show that DIAMOND dynamics satisfies (\ref{eq:definition_exact}) with the following exact potential function:
\begin{equation}
    \phi(\sigma) = \sum_{n=1}^{N} u_{n}(\sigma), \label{eq:exact_potential}
\end{equation}
namely, the overall objective function that needs to be maximized. To show this, consider two strategies for flow $n_0$: $\sigma_{n_0}^{(1)}, \sigma_{n_0}^{(2)}\in\mathcal{A}_n$,  and fix the strategy profile for all other flows $\sigma_{-n_0}$. Then, the difference in the potential function \eqref{eq:exact_potential} is given by:
\begin{equation}
\begin{array}{l}
\displaystyle
\Delta\phi(\sigma_{n_0}^{(1)}, \sigma_{n_0}^{(2)}, \sigma_{-n_0})\vspace{0.2cm}\\
    \displaystyle
\triangleq 
    \phi(\sigma_{n_0}^{(2)}, \sigma_{-n_0}) - \phi(\sigma_{n_0}^{(1)}, \sigma_{-n_0})\vspace{0.2cm}\\
    \displaystyle
= 
    \left(u_{n_{0}}(\sigma_{n_0}^{(2)}, \sigma_{-n_0}) + \sum_{m\neq {n_0}} u_{m}(\sigma_{n_0}^{(2)}, \sigma_{-n_0})\right)\vspace{0.2cm}\\
    \displaystyle
    -\left(u_{n_{0}}(\sigma_{n_0}^{(1)}, \sigma_{-n_0}) + \sum_{m\neq {n_0}} u_{m}(\sigma_{n_0}^{(1)}, \sigma_{-n_0})\right)\vspace{0.2cm}\\
    \displaystyle
= 
    \left(u_{n_{0}}(\sigma_{n_0}^{(2)}, \sigma_{-n_0}) + \sum_{m\in \mathcal{N}(n_0)} u_{m}(\sigma_{n_0}^{(2)}, \sigma_{-n_0})\right)\vspace{0.2cm}\\
    \displaystyle
    -\left(u_{n_{0}}(\sigma_{n_0}^{(1)}, \sigma_{-n_0}) + \sum_{m\in\mathcal{N}(n_0)} u_{m}(\sigma_{n_0}^{(1)}, \sigma_{-n_0})\right)\vspace{0.2cm}\\
    \displaystyle
=\mathcal{U}_{n_0}(\sigma_{n_0}^{(2)}, \sigma_{-n_0}) -\mathcal{U}_{n_0}(\sigma_{n_0}^{(1)}, \sigma_{-n_0})\vspace{0.2cm}\\
\displaystyle
\triangleq
\Delta\mathcal{U}_{n_0}(\sigma^{(1)}, \sigma^{(2)}, \sigma_{-n_0}),
\end{array}    
\end{equation}
where the second equality follows since flow $n_0$ only interferes with its neighbors $\mathcal{N}(n_0)$. As a result, their utility might change, while the utilities of all other flows $\mathcal{F} \setminus \mathcal{N}(n_0)$ remain the same and are cancelled. Furthermore, since the utility $u_{n_0}$ is bounded by $u_{max}$, it follows that $\phi(\sigma)<N\cdot u_{max} < \infty$ holds. As a result, $\phi(\sigma)$ in \eqref{eq:exact_potential} is a bounded exact potential function of the game, which proves that DIAMOND dynamics is an exact potential game.

Finally, recall that DIAMOND plays noisy BR dynamics by the NB3R module with respect to $\mathcal{U}_n$. We now use the fact that a noisy BR dynamics following (\ref{eq:nb3r_pmf})
in the exact potential game converges to a stationary
distribution of the Markov chain corresponding to the game \cite{blume1993statistical}. Since DIAMOND is an exact potential game with an exact potential function $\phi$ given in (\ref{eq:exact_potential}), the stationary distribution of the strategy profile is given by:
 \begin{equation}
\displaystyle\mathbb{P}(\sigma^{DIAMOND(\nu)}=\sigma) = \displaystyle\frac{e^{\nu\cdot\phi(\sigma)}}{\sum_{\tilde{\sigma}} e^{\nu\cdot\phi(\tilde{\sigma})}}. 
\label{eq:prob_sigma}
 \end{equation}
Note that the strategy that maximizes \eqref{eq:exact_potential} is the one that solves  \eqref{eq:objective} as well. Thus, its global maximum lies inside the action space played by the NB3R module. Therefore, for any $\epsilon>0$ we can choose $\nu>0$ sufficiently large such that the stationary distribution puts
a sufficiently high weight on the strategy profile that maximizes \eqref{eq:exact_potential} (i.e., $\phi$ in \eqref{eq:prob_sigma}). Thus, \eqref{eq:prob_nb3r} is satisfied as time approaches infinity.
$\qed$ \vspace{0.2cm}

\emph{Corollary 1:} Let $\sum_{n=1}^{N} u_n\left(\sigma^{DIAMOND\left(\nu(t)\right)}\right)$ and  
$\sum_{n=1}^{N} u_n\left(\sigma^*\right)$ be the sum utility achieved by DIAMOND at time $t$ and the optimal strategy, respectively. Let $\Delta=N\cdot u_{max}$, and let the parameter $\nu$ in \eqref{eq:nb3r_pmf} increase with time as $\nu(t)=\log (t)/\Delta$. Then,
\begin{equation}
\begin{array}{l}
\displaystyle
\sum_{n=1}^{N} u_n\left(\sigma^{DIAMOND\left(\nu(t)\right)}\right)\longrightarrow\sum_{n=1}^{N} u_n\left(\sigma^*\right) \vspace{0.2cm}\\\hspace{6cm}
\mbox{as\;\;}t\longrightarrow\infty.\vspace{0.2cm}    
\end{array}
\end{equation}

\emph{Proof:}
Following the proof of Theorem \ref{th:main_theorem}, the stationary distribution of the Markov chain with a fixed $\nu$ is given by \eqref{eq:prob_sigma}. As a result, as the probability of playing BR increases, where BR is the strategy of each flow maximizing the improvement in the collaborative utility at each updating time given the current state of other flows (i.e., by increasing $\nu$), the probability of attaining the global maximum of the potential function \eqref{eq:exact_potential} increases with time. 
Achieving the optimal solution (i.e., letting $\epsilon\rightarrow 0$ in \eqref{eq:prob_nb3r}) requires $\nu$ to approach infinity. However, increasing $\nu$ too fast may push the algorithm into a local maximum for a long time (since the probability of not playing BR is too small). The process of increasing $\nu(t)$ during the algorithm is also known as cooling the system in simulated annealing analysis, where $1/\nu(t)$ represents the temperature. Following simulated annealing analysis \cite{hajek1988cooling}, convergence to the optimal solution is attained by increasing $\nu(t)$ as $\nu(t)=\log t/\Delta, t=1, 2, ...$, where it suffices to set the constant $\Delta$ to be greater than the maximal change in the objective function, which is upper bounded by $N\cdot u_{max}$ in our case. $\qed$\vspace{0.2cm} 

Using the DIAMOND dynamics described in Corollary 1, the flows explore strategy profiles at the beginning of the algorithm, and converge to their best response as time approaches infinity. In cases where multiple optimal operating points exist, the algorithm may converge to one of them. However, simulation results show that the optimal solution can be reached much faster with smaller values of $\Delta$ under typical scenarios.

Note that in order to handle the vast search space of the multi-flow interference-aware transmission problem, DIAMOND employs a strategic approach by constructing $K_n$ distinct routes for each flow $n$ using Algorithm \ref{alg:k_paths}. Setting a sufficiently high value for $K_n$ could potentially result in $\mathcal{A}_n$ containing all possible paths between the source $s_n$ and destination $d_n$ for each flow $n$. However, such an approach would significantly enlarge the search space, rendering the problem computationally challenging to solve efficiently. Consequently, it is common to restrict the search space due to the combinatorial nature of the problem. The DIAMOND algorithm aims to find the optimal solution within the restricted search space, but it does not provide a guarantee of achieving the optimal solution under an unrestricted search space.

\subsection{Numerical Analysis}
\label{ssec:numerical}

We now provide numerical examples to illustrate the performance of the proposed DIAMOND algorithm in various wireless interference network environments.

In our simulations, all $V$ network nodes were randomly deployed within an $A \times A$-sized square area, were $A$ was set to $1000$m. In a random topology, $E$ links were generated randomly to ensure that the graph forms a single connected component. The flow demands (source, destination, and packet load) were also randomly generated. For NFSNET and GEANT2, the link connectivity was not randomized in order to maintain the structure of the graph. 

The GRRL agent was initially trained on a small-scale problem with parameters: $N=20, V=10, E=20$, and subsequently fine-tuned on a larger-scale setting with parameters: $N=30, V=20, E=30$. To highlight the generalization capabilities of the GNN and the adaptive NB3R algorithm, all testing results were obtained from different unseen problem settings, demonstrating the generalization ability of our method to new topologies and varying flow demand.

We compared the following algorithms: (i) \emph{Random Baseline (RB)}: The random baseline is a heuristic method that chooses the best path allocation out of $100$ independent trails of random choices from the action set. A random baseline is commonly used in machine learning research to evaluate the ability of a learning algorithm to learn good strategies and perform better than exhaustive search over a limited number of trials. (ii) \emph{Open Shortest Path First (OSPF)} \cite{OSPF}: The OSPF protocol (and its variants) is widely used for routing data and is implemented in many real-world systems. OSPF uses Dijkstra algorithm to compute and transmit data through the shortest path. (iii) DQN+GNN \cite{DRL+GNN}: The recently suggested DQN+GNN algorithm uses a DRL algorithm, based on the well-known offline-RL DQN \cite{mnih2015human} algorithm. (iv) \emph{Delay-and Interference-Aware Routing (DIAR)} \cite{Chai2020DIAR}: In a recently suggested DIAR algorithm, a centralized unit selects routes for all flows. DIAR aims to select routes with minimal end-to-end delay for multiple concurrent data flows. It employs an improved genetic algorithm (GA) to balance the network load and minimize delay. (v) \emph{Interference Aware Cooperative Routing (IACR)} \cite{Waqas2022IACR}: IACR was introduced recently as a distributed routing method for 5G networks. It selects routes for flow demands based on a weighted sum of created and received interference in the network. In the implementation of all algorithms, we used the same knowledge of the wireless network state and interference map to optimize the objective and compute the selected paths. We used a bandwidth of $20$Mhz for all links. For each simulation experiment, we averaged the results over $10$ random independent settings, i.e., flow demands and link capacities. 

\begin{figure*}
\begin{center}
    \subfigure[Random Topology $V=60, E=90$]{\scalebox{0.31}
    {
      \label{fig:rand6090_topology_a}
      \includegraphics{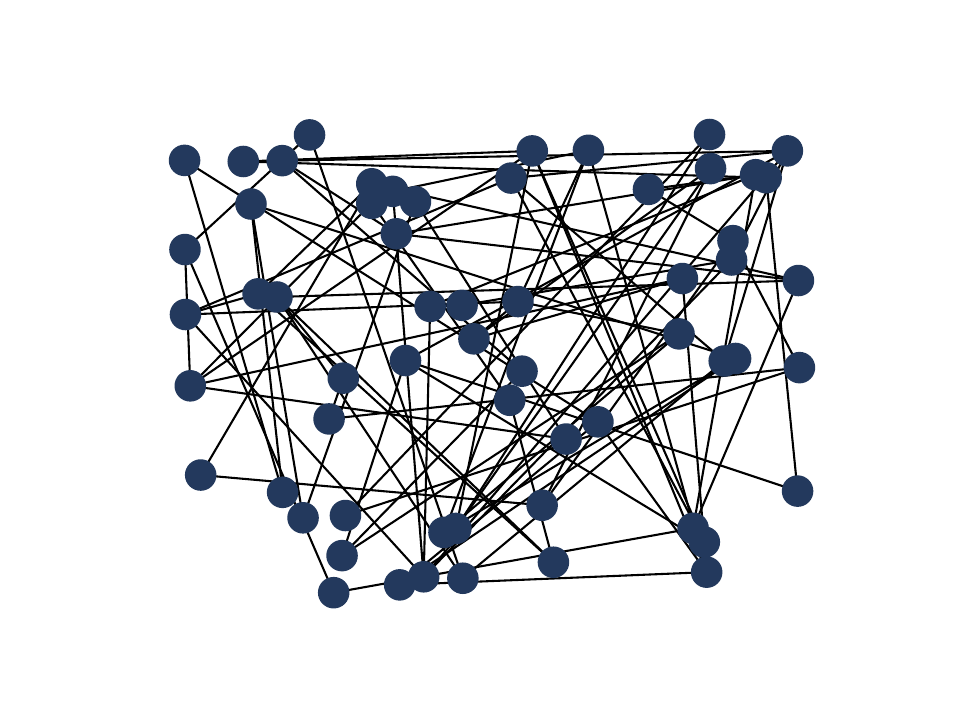}
    }}
    \hspace{-1cm}\hfill
    \subfigure[Average Flows Rates]{\scalebox{0.3}
    {
    \label{fig:rand6090_rate}
    \input{figures/V60E90/Rates.tex}
    }}
    \hfill
    \subfigure[Network Delay]{\scalebox{0.3}
    {
        \label{fig:rand6090_delay_c}
      \input{figures/V60E90/Delay.tex}
    }}
    \caption{Performance evaluation of the algorithms as a function of the number of flows in a large-size dense cluster network.}
\end{center}
\end{figure*}

To compute the average flow rate, we follow the procedure outlined next: Consider flow $n$ with a route that starts at the source node $s_n$ and passes through nodes $v_1, v_2, ..., v_{D}$ before reaching the destination node $d_n$. For each link along this route, denoted as $(s_n, v_1)$, $(v_1, v_2)$, ..., $(v_D, d_n)$ (where $(v, v')$ represents a link between node $v$ and node $v'$), we calculate its capacity using the Shannon formula: $C_{\ell}=B_{\ell}\log_2(1+\frac{P_{\ell}|h_{\ell}|^2}{\tilde{\sigma}^2+I_{\ell}})$, where $B_{\ell}, P_{\ell}, |h_{\ell}|, I_{\ell}, \tilde{\sigma}^2$ denote the bandwidth, transmission power, channel gain, interference power (i.e., received power from interfering nodes, computed based on the interference map), and the noise power at link $\ell$, respectively. Each link utilizes TDMA for equal-time sharing among all flows that pass through it (say $M_{\ell}$ flows pass through link $\ell$). Then, the available capacity for flow $n$ at link $\ell$ is given by $C_n(\ell)=\frac{1}{M_{\ell}}C_{\ell}$. Consequently, the rate of flow $n$ is determined by the link along its route with the minimal capacity: $R_n=\min_{\ell\in \varphi^n}C_n(\ell)$, where $\varphi^n=\left((s_n, v_1), (v_1, v_2), ..., v_D, d_n)\right)$ denotes the route of flow $n$. We average this rate over all flows to obtain the average flow rate. Additionally, we computed the delay as follows. For all flows, for each packet inserted to the system we count the total number of time steps until reaching to destination. Each hop (i.e., link) along the route is considered as one time step, and we also account for the time spent in the link queues by adding it to the overall count. We average this counter over all packets to obtain the average delay. 
Specifically, for each packet we count the total number of time steps from when a packet is initially inserted into the source node until it reaches the destination node. Consider a packet, denoted as $p$, traversing a route $r=(n_1, n_2, ... n_d)$, where $n_1$ denotes the source node and $n_d$ denotes the destination node. Let $D_i$ represent the delay experienced by packet $p$ across the link $(n_i,n_{i+1})$. The total delay endured by packet $p$ is given by the summation $\sum_{i=1}^{d-1}D_i$. Each $D_i$ comprises two components: The count of time steps the packet waits within the queue (queue delay), along with an additional time step for transmission (transmission delay). The rate at which packets are serviced hinges on the link rate, considering solely the link rate corresponding to the flow's direction in a full-duplex mode. We point out that we disregard the propagation delay, which contributes a consistent delay contingent on link distance and signal speed through the medium. As a result, it becomes less significant when comparing different routing algorithmic methods. Also, in many instances, this component can be disregarded as in local area networks, data centers, and inter-system communication, characterized by relatively short distances. Consequently, we exclude this element from our simulations (in all tested methods to ensure a fair comparison). Similar computations have been employed in previous works 
\cite{Ju2012fullduplex, Stamatiou2010delayopt, Li2018clustering, Vondrous2016performance, Bisnik2006delay}.

In the simulations, we set the utility of flow $n$ to $u_n(\sigma) = R_n(\sigma)$ (i.e., its achievable rate). DIAMOND is designed to support both half-duplex and full-duplex transmissions. Our simulations assume the use of the full-duplex mode. When multiple flows are being transmitted, the transmitting nodes employ the TDMA access protocol and utilize RTS/CTS signals prior to transmission.\\

First, we simulated a large-size dense cluster. We simulated a random network setting, where each trial involves randomizing $N=60$ node positions, $E=90$ random links between nodes, as well as random capacities and random flow demands. For convenience, we refer to the case of $N\leq 0.5V$ as a lightly-loaded network, the case of $0.5V < N\leq V$ as a moderately-loaded network, and the case of $N>V$ as a heavily-loaded network. The results are presented in Figs. \ref{fig:rand6090_topology_a}-\ref{fig:rand6090_delay_c}. In all cases, DIAMOND shows significantly better performance than all other algorithms, with an average flow rate (Fig. \ref{fig:rand6090_rate}) outperforming DIAR (in second place) by $20\%$ to $70\%$ in lightly-loaded networks. The performance gain increases to $100\%$ or more as the load increases. It is important to notice that the superior performance of DIAMOND in achievable rate does not come at a cost of the packet delay. On the contrary, it can be seen in Fig. \ref{fig:rand6090_delay_c} that DIAMOND significantly outperforms all other algorithms in terms of packet delay as well for most network load values.

\begin{figure*} 
\begin{center}
    \subfigure[NSFNET \cite{NSFNET}]{\scalebox{0.31}
    {
    \label{fig:nsfnet_topology1}
      \includegraphics{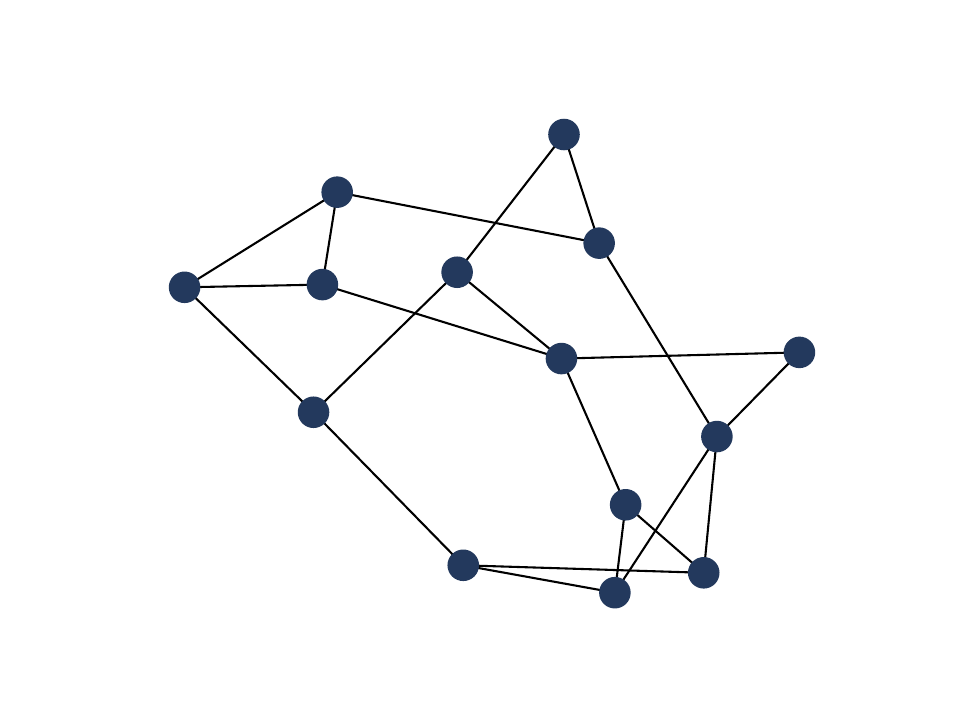}
    }}
    \hspace{-1cm}\hfill
    \subfigure[Average Flows Rates]{\scalebox{0.3}
    {
    \label{fig:nsfnet_rate1}  
    \input{figures/NSFNET/Rates.tex}
    }}
    \hfill
    \subfigure[Network Delay]{\scalebox{0.3}
    {
    \label{fig:nsfnet_delay1}
      \input{figures/NSFNET/Delay.tex}
    }}
    \caption{Performance evaluation of the algorithms as a function of the number of flows in NSFNET network.}
\label{fig:performance_topologies_NSFNET}
\end{center}
\end{figure*}

Second, we evaluate the performance over NSFNET topology, which is a known topology with $V=14$ nodes and $E=21$ edges \cite{NSFNET}. This experiment corresponds to a small-size cluster of users. We randomize over trials the node positions, link capacities and flow demands. The results are presented in Figs. \ref{fig:nsfnet_topology1}-\ref{fig:nsfnet_delay1}. In all cases, DIAMOND shows significantly better performance than all other algorithms, with an average flow rate (Fig. \ref{fig:nsfnet_rate1}) outperforming DIAR (in second place) by $10\%$ to $40\%$. It is important to note again that the superior performance of DIAMOND in achievable rate does not come at the cost of packet delay. As seen in Fig. \ref{fig:nsfnet_delay1}, DIAMOND outperforms all other algorithms in terms of packet delay for most network load values. 

\begin{figure*} 
\begin{center}
    \subfigure[GEANT2 \cite{GEANT2}]{\scalebox{0.31}
    {
    \label{fig:geant2}
      \includegraphics{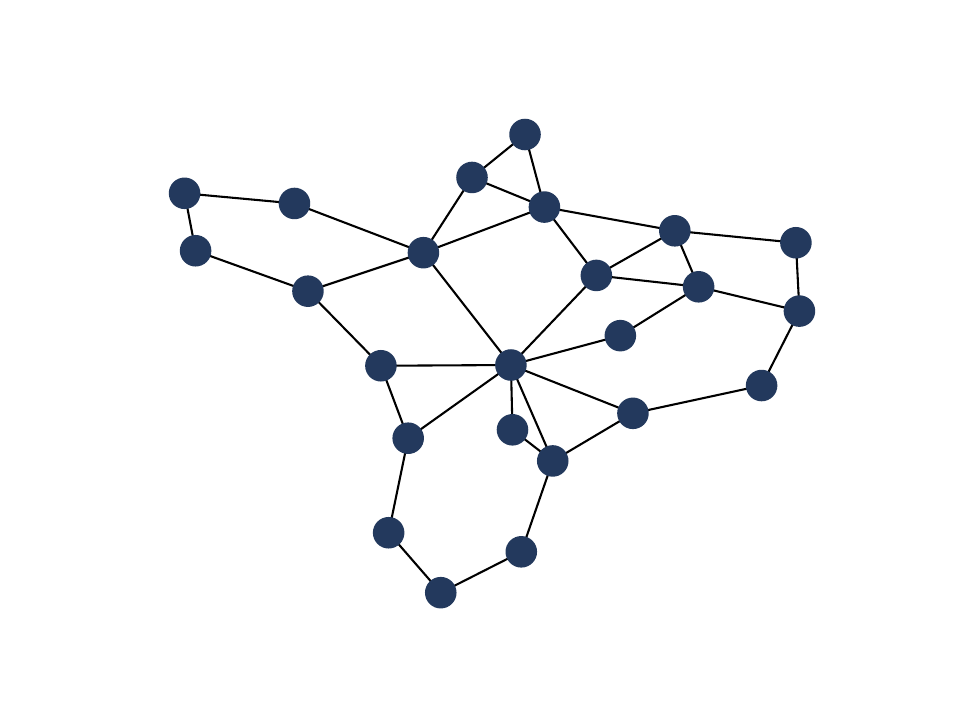}
    }}
    \hspace{-1cm}\hfill
    \subfigure[Average Flows Rates]{\scalebox{0.3}
    {
    \label{fig:geant2_rate}  
    \input{figures/Geant2/Rates.tex}
    }}
    \hfill
    \subfigure[Network Delay]{\scalebox{0.3}
    {
    \label{fig:geant2_delay}
      \input{figures/Geant2/Delay.tex}
    }}
    
    \caption{Performance evaluation of the algorithms as a function of the number of flows in GEANT2 network.}
\label{fig:performance_topologies_GEANT2}
\end{center}
\end{figure*}

Third, we evaluate the performance over GEANT2 topology, which is a known topology with $V=24$ nodes and $E=37$ edges \cite{GEANT2}. This experiment corresponds to a medium-size cluster of users. We randomize over trials the node positions, link capacities and flow demands. The results are presented in Figs. \ref{fig:geant2}-\ref{fig:geant2_delay}. In terms of average flow, DIAMOND shows significantly better performance than all other algorithms, with an average flow rate (Fig. \ref{fig:geant2_rate}) outperforming DIAR (in second place) by $10\%$ to $40\%$. As seen in Fig. \ref{fig:geant2_delay}, DIAMOND achieves good performance as compared to the other algorithms in terms of packet delay as well.

Note that Figs. \ref{fig:performance_topologies_NSFNET}, \ref{fig:performance_topologies_GEANT2} present simulation results for a specific small network topology (NFSNET and GEANT2, respectively). While DIAMOND generally achieves the best results in most simulated cases, it is worth acknowledging that due to the inherent randomness of learning algorithms, slight performance degradation can occur at certain points, even for strong algorithms, and even when averaging over different trials. This phenomenon has been commonly observed in the field of RL literature, as demonstrated in e.g., \cite{jiang2020dgn, kortvelesy2022qgnn, liu2020pic, van2016ddqn}. However, despite any minor performance variations caused by noise, DIAMOND consistently achieves higher average flow rates compared to other algorithms, with only moderate decreases in terms of delay. Overall, the proposed method demonstrates superior performance compared to its competitors even in this specific network topology.
In contrast, in the case of the large random topology, where the averaging of results over multiple random topologies can mitigate the impact of noise, DIAMOND exhibits significantly better performance across all simulation points. The random topology setting better represents the broader real-world implementation scenario due to the inherent randomness of wireless networks.

We also note that communication networks typically retain the ability to effectively support high numbers of flows, even when surpassing the count of nodes. We next explain the reason for this. Consider the scenario where a specific link, denoted as $\ell$, is operating at full capacity. Denote the number of flows passing through link $\ell$ as $M_{\ell}$. Given that each link employs TDMA to ensure equal-time allocation among all traversing flows, the aggregate link rate becomes distributed among these $M_{\ell}$ flows. Concurrently, the interference time attributed to each individual flow is divided by the factor $M_{\ell}$. The interference stemming from link $\ell$ remains bounded by the maximum permissible transmission power, multiplied by the interference channel power. This holds true even when the link is fully utilized, irrespective of the volume of flows passing through it. The exact numerical value is contingent upon variables such as the link's capacity and the attributes of the interference channel, aligning in the transmission direction within a full-duplex mode. This observation is demonstrated in Fig. \ref{fig:performance_topologies_NSFNET}.

\begin{figure*} 
\begin{center}
    \subfigure[Random Topology $V=30, E=50$]{\scalebox{0.32}
    {
      \label{fig:rand3050_topology}
      \includegraphics{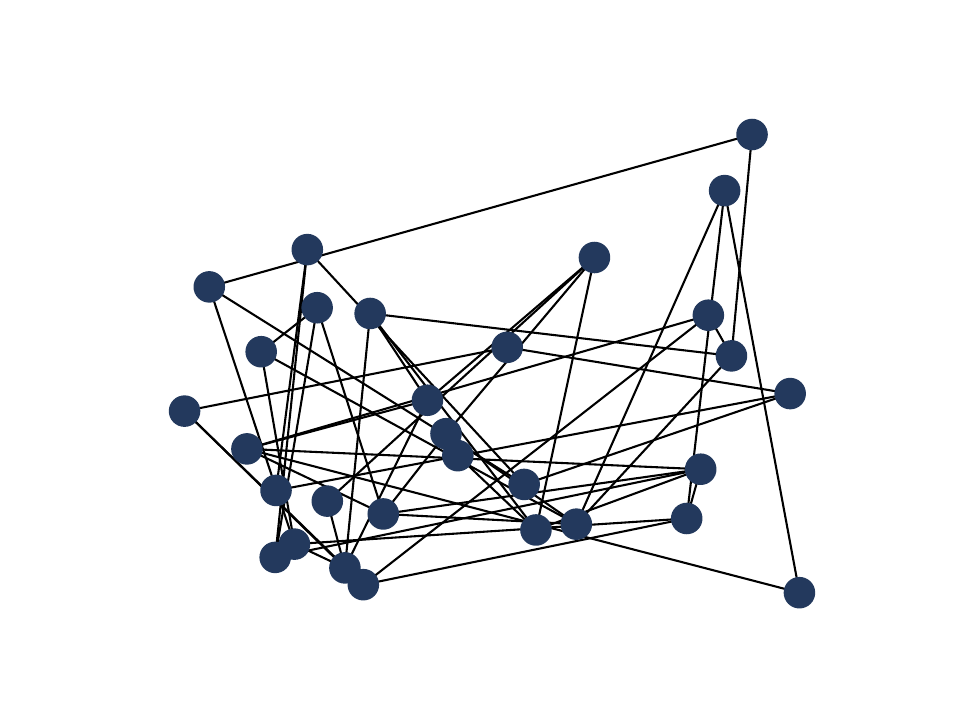}
    }}
    
    \hfill
    \subfigure[Average Flow Rates]{\scalebox{0.3}
    {
    \label{fig:rand3050_rate}      \input{figures/V30E50/Rates_equal.tex}
    }}
    \hfill
    \subfigure[Network Delay]{\scalebox{0.3}
    {
        \label{fig:rand3050_delay}
      \input{figures/V30E50/Delay_equal.tex}
    }}\hspace{2.5cm}

    \hfill
    \subfigure[Average Flow Rates]{\scalebox{0.3}
    {
    \label{fig:rand3050_rate_steps}      \input{figures/V30E50/Rates_steps.tex}
    }}
    \hfill
    \subfigure[Network Delay]{\scalebox{0.3}
    {
        \label{fig:rand3050_delay_steps}
      \input{figures/V30E50/Delay_steps.tex}
    }}\hspace{2.5cm}

    \hfill
    \subfigure[Average Flow Rates]{\scalebox{0.3}
    {
    \label{fig:rand3050_rate_rayleigh}      \input{figures/V30E50/Rates_rayleigh.tex}
    }}
    \hfill
    \subfigure[Network Delay]{\scalebox{0.3}
    {
        \label{fig:rand3050_delay_rayleigh}
      \input{figures/V30E50/Delay_rayleigh.tex}
    }}\hspace{2.5cm}
    \caption{Performance evaluation of the algorithms in a randomly deployed network consisting of $V=30$ nodes and $E=50$ links using different transmission power distributions. The top subplot represents the performance using the constant transmission power. The middle subplot illustrates the performance using the path loss-aware transmission power level. The bottom subplot shows the performance using the channel-aware power control.}
\label{fig:rand3050_power}
\end{center}
\end{figure*}

\begin{table*}[ht]
	\caption{Algorithm comparison for various network configurations. 
}
	\label{tab:rand_performances}
	\begin{center}
		\scalebox{1}{
			\begin{tabular}{c c c || c c c c | c c c c}
                \toprule
                & & & \multicolumn{4}{c|}{Average Flow Rates [Mbps] $\uparrow$} & \multicolumn{4}{c}{Max Delay [time-steps] $\downarrow$}\\
				\midrule 
				$N$ & $V$ & $E$ & RB & OSPF & DQN+GNN & \textbf{DIAMOND (ours)} & RB & OSPF & DQN+GNN & \textbf{DIAMOND (ours)}  \\ \hline
    
				\hypertarget{r1}{}25 & 50 & 75 & 25.624 & 26.78 & 46.636 &  \textbf{80.216} & 31.3 & 72.0 & 70.4 &  \textbf{16.5}\\ 

                \hypertarget{r2}{}100 & 200 & 300 & 0.611 & 2.061 & 9.937 &  \textbf{17.239} & 314.2 & 372.1 & 339.9 &  \textbf{113.8}\\

                \hypertarget{r3}{}70 & 70 & 140 & 1.887 & 4.201 & 20.597 &  \textbf{30.594} & 179.2 & 147.7 & 155.5 &  \textbf{60.7}\\

                \hypertarget{r4}{}20 & 20 & 40 & 26.935 & 60.77 & 75.505 &  \textbf{108.14} & 55.2 & 10.8 & 35.6 &  \textbf{8.4}\\

                \hypertarget{r5}{}80 & 40 & 60 & 0.57625 & 1.841 & 9.85 &  \textbf{18.107} & 192.0 & 191.7 & 182.3 & \textbf{130.6}\\

                \hypertarget{r6}{}100 & 50 & 75 & 0.276 & 0.841 & 6.468 & \textbf{13.6} & 311.4 & 343.8 & 419.1 & \textbf{190.8}\\

                \hypertarget{r7}{}50 & 10 & 15 & 6.504 & 9.05 & 25.024 & \textbf{41.054} & 40.8 & 36.8 & 49.2 & \textbf{24.7}\\

                \hypertarget{r8}{}150 & 30 & 45 & 0.1113 & 0.285 & 4.026 & \textbf{7.894} & 544.6 & 684.5 & 577.2 & \textbf{195.2}\\

                \hypertarget{r9}{}140 & 70 & 105 & 0.1207 & 0.396 & 4.137 & \textbf{8.89} & 544.5 & 608.7 & 513.2 & \textbf{455.3}\\

                \hypertarget{r10}{}30 & 10 & 45 & 18.553 & 20.21 & 52.57 & \textbf{77.689} & 18.8 & 24.2 & 26.1 & \textbf{10.8}\\

            \bottomrule
			\end{tabular}
		}
	\end{center}
\end{table*}

\begin{figure*}[h] 
\begin{center}
    \subfigure[Random topology $V=50, E=80$]{\scalebox{0.31}
    {
      \label{fig:rand5080_topology}
      \includegraphics{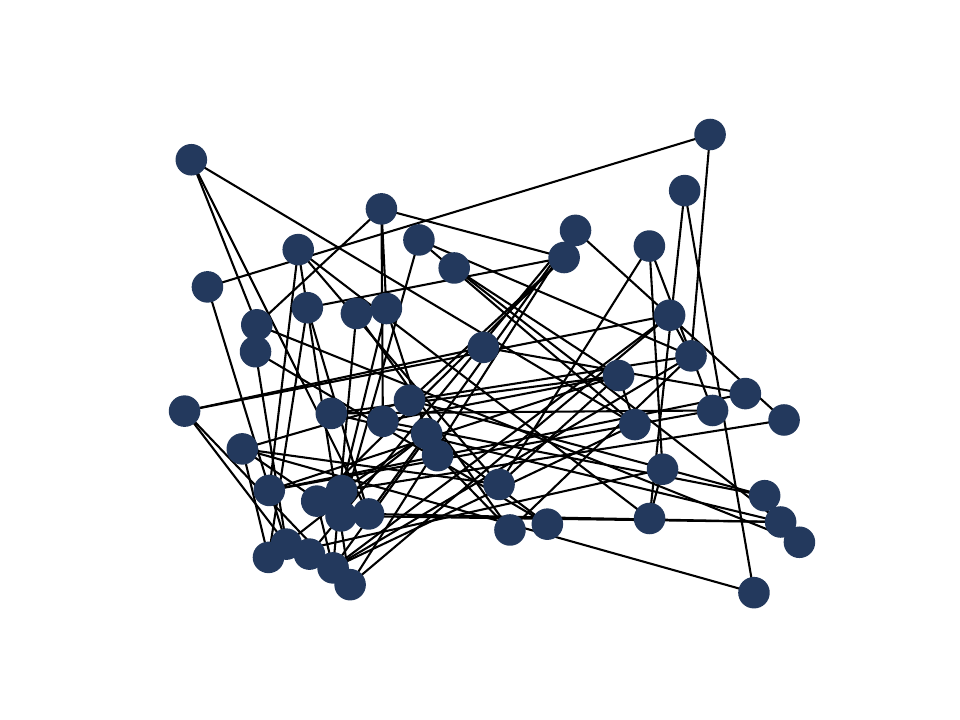}
    }}
    \subfigure[Average Flow Rates]{\scalebox{0.3}
    {
      \input{figures/NB3R/NB3RRates.tex}
    }}
    \hfill
    \subfigure[Network delay]{\scalebox{0.3}
    {
      \input{figures/NB3R/NB3RDelay}
    }}
    \caption{Convergence analysis achieved by DIAMOND algorithm over a random topology network with parameters: $N=20, V=50, E=80$.}
  \label{fig:NB3R convergance}
\end{center}
\end{figure*}

Forth, we have simulated three different commonly used power distributions: 
(i) \emph{Constant transmission power} (Figs. \ref{fig:rand3050_rate}-\ref{fig:rand3050_delay}): In this case, the transmission power of all links (say $P_\ell$ at link $l$) is set to $P_\ell = P$ for all active links $\ell \in \mathcal{E}$. In the simulations, we have normalized the transmission power to $P=1$.
(ii) \emph{Path loss-aware transmission power level} (Figs. \ref{fig:rand3050_rate_steps}-\ref{fig:rand3050_delay_steps}): In this case, each transmission at each link utilizes different power level for transmission (e.g., low, mid, and high, as used in the simulations). The transmission power is adapted based on the link distance, so that links with higher distance (i.e., higher free-space path loss over the link) use higher power level for transmission. In the simulations, the nodes were deployed randomly within an $A \times A$-sized square area (i.e., the maximal distance was $\bar{L} = \sqrt{2}A$). In this case, we set the transmission power $P_\ell$ for each active link $\ell$ as follows: 
$P_\ell=P_{\textrm{high}}$, if the link distance is greater or equal to $\frac{2 \bar{L}}{3}$, $P_\ell=P_{\textrm{mid}}$, if the link distance is greater or equal to $\frac{ \bar{L}}{3}$ and smaller than $\frac{2 \bar{L}}{3}$, $P_\ell=P_{\textrm{low}}$ if the link distance is smaller than $\frac{ \bar{L}}{3}$. We set $P_{\textrm{high}}=1, P_{\textrm{mid}}=2/3, P_{\textrm{low}}=1/3$. 
(iii) \emph{Channel-aware power control} (Figs. \ref{fig:rand3050_rate_rayleigh}-\ref{fig:rand3050_delay_rayleigh}): In this case, each transmission utilizes its CSI to cancel the fading effect by amplifying the signal by its inverse channel coefficient. However, weak channels might cause an arbitrarily high amplification, possibly violating a transmission power constraint (say $P_{max}$). Therefore, the maximal power is limited by $P_{max}$. Therefore, we set $P_\ell = \max(P_{max}, P/h_l)$ for all active links $\ell \in \mathcal{E}$. In the simulations, we simulated Rayleigh fading channels over links.\\
We evaluated the performance of the algorithms on a randomly deployed network comprising $30$ nodes and $50$ links $(V = 30, E = 50)$. We conducted three experiments, one for each power distribution as explained above. 

Fig. \ref{fig:rand3050_power} illustrates the robustness of our proposed DIAMOND algorithm when subjected to different power distribution transmissions. It is noteworthy that DIAMOND was solely trained on an equal normalized power distribution, yet it demonstrates the ability to generalize and effectively capture the global network state across various scenarios, even when applied to an unseen network topology during training.

To further validate the robustness of DIAMOND in supporting various network configurations with changing numbers of flow demands ($N$), nodes ($V$), and edges ($E$), and to justify its usage in 5G topologies, which consist of dense networks with a large number of edges, we conducted additional experiments using randomized settings. The results are presented in Table \ref{tab:rand_performances}.
First, we simulated lightly-loaded networks where $V=2N$ and $E=1.5V$ (lines \hyperlink{r1}{1}, \hyperlink{r2}{2}). Second, we simulated moderately-loaded networks where $V=N$ and $E=2V$ (lines \hyperlink{r3}{3}, \hyperlink{r4}{4}). Third, we simulated heavily-loaded networks where $V=0.5N$ and $E=1.5V$ (lines \hyperlink{r5}{5}, \hyperlink{r6}{6}) and $V=0.2N$ and $E=1.5V$ (lines \hyperlink{r7}{7}, \hyperlink{r8}{8}). Lastly, we considered a scenario with a higher and lower ratio between nodes and edges (lines \hyperlink{r9}{9}, \hyperlink{r10}{10}, respectively). The results in Table \ref{tab:rand_performances} show that DIAMOND significantly outperforms all other algorithms in terms of both average flow rate and packet delay in all scenarios, which represent different realistic network configurations.

Table \ref{tab:rand_performances} demonstrates the generalization performance of our GNN architecture across a large number of cases. It covers different scenarios involving network configurations with varying sizes and nodes-links ratios, as well as different numbers of flow demands to nodes ratios. These results collectively demonstrate that our proposed method is capable of effectively generalizing to unseen scenarios and exhibits strong performance across a range of network configurations and flow demands.

Next, we show the convergence of NB3R in a representative setting. We evaluated the performance of the algorithms on a randomly deployed network comprising $50$ nodes, $80$ links and $20$ flows $(N=20, V = 50, E = 80)$, for finite updates duration  within $\tilde{t}=1,2,\ldots,100$ time-slots.

Fig. \ref{fig:NB3R convergance} demonstrates numerically the convergence of DIAMOND in a representative setting of a random topology network with parameters: $N=20, V=50, E=80$. The results were averaged over $100$ i.i.d. trails. Figures (b) and (c) show the increase in average flow rate, and the decrease in network delay, respectively, with time until convergence.

In summary, the simulation results demonstrate that DIAMOND consistently exhibits very strong performance when compared to all other algorithms.

\section{Conclusion}
\label{sec:conclusion}

The problem of multi-flow transmission in wireless networks, where mutual interference between links can reduce link capacities, has been addressed with the development of a novel algorithm called DIAMOND. The algorithm allows for a hybrid centralized-distributed implementation and is designed to maximize the network utility. The theoretical analysis has proven that DIAMOND converges to the optimal multi-flow transmission strategy over time. Extensive simulation results over various network topologies have demonstrated the superior performance of DIAMOND compared to existing methods. DIAMOND represents a promising approach to improving multi-flow transmission in wireless networks and has the potential to make valuable contributions in 5G and beyond technologies with centralized unit deployments.

\bibliographystyle{IEEEtran}
\bibliography{main}

\end{document}

%% file: figures/V60E90/Rates.tex
\pgfplotstableread[col sep=comma]{figures/V60E90/random_rates_V60_E90_all_comps.csv}\data
\begin{tikzpicture}
\begin{axis}[
    width=\textwidth,
    height=10cm,
    xtick=data,
    xticklabels from table={\data}{N},
    legend pos= north east,
    legend cell align={left},
    ylabel={Avg. Flows Rate [Mbps]},
    xlabel={Number of flows $N$},
    grid=both,
    grid style=dashed,
    ]
    
    \addplot [mark=o, blue] table [x expr=\coordindex, y={DIAMOND}]{\data};
    \addlegendentry{DIAMOND (ours)}

    \addplot [mark=square, orange] table [x expr=\coordindex, y={DQN+GNN}]{\data};
    \addlegendentry{DQN+GNN}

    \addplot [mark=pentagon, cyan] table [x expr=\coordindex, y={DIAR}]{\data};
\addlegendentry{DIAR}

    \addplot [mark=star, black] table [x expr=\coordindex, y={IACR}]{\data};
    \addlegendentry{IACR}
    
    \addplot [mark=diamond, green!40!gray] table [x expr=\coordindex, y={OSPF}]{\data};
    \addlegendentry{OSPF}
    
    \addplot [mark=triangle, violet] table [x expr=\coordindex, y={RandomBL}]{\data};
    \addlegendentry{RB}
    
\end{axis}
\end{tikzpicture}

%% file: figures/V60E90/Delay.tex
\pgfplotstableread[col sep=comma]{figures/V60E90/random_delay_V60_E90_all_comps.csv}\data

\begin{tikzpicture}
\begin{axis}[
    width=\textwidth,
    height=10cm,
    xtick=data,
    xticklabels from table={\data}{N},
    legend pos=north west,
    legend cell align={left},
    ylabel={Delay [timesteps]},
    xlabel={Number of flows $N$},
    grid=both,
    grid style=dashed,
    ]
    
    \addplot [mark=triangle, violet] table [x expr=\coordindex, y={RandomBL}]{\data};
    \addlegendentry{RB}

    \addplot [mark=square, orange] table [x expr=\coordindex, y={DQN+GNN}]{\data};
    \addlegendentry{DQN+GNN}

    \addplot [mark=diamond, green!40!gray] table [x expr=\coordindex, y={OSPF}]{\data};
    \addlegendentry{OSPF}

    \addplot [mark=pentagon, cyan] table [x expr=\coordindex, y={DIAR}]{\data};
    \addlegendentry{DIAR}

    \addplot [mark=star, black] table [x expr=\coordindex, y={IACR}]{\data};
    \addlegendentry{IACR}

    \addplot [mark=o, blue] table [x expr=\coordindex, y={DIAMOND}]{\data};
    \addlegendentry{DIAMOND (ours)}
    
\end{axis}
\end{tikzpicture}

%% file: figures/NSFNET/Rates.tex
\pgfplotstableread[col sep=comma]{figures/NSFNET/nsfnet_rates_all_comps.csv}\data

\begin{tikzpicture}
\begin{axis}[
    width=\textwidth,
    height=10cm,
    xtick=data,
    xticklabels from table={\data}{N},
    legend pos=north east, 
    legend cell align={left},
    ylabel={Avg. Flows Rate [Mbps]},
    xlabel={Number of flows $N$},
    grid=both,
    grid style=dashed,
    ]
    
    \addplot [mark=o, blue] table [x expr=\coordindex, y={DIAMOND}]{\data};
    \addlegendentry{DIAMOND (ours)}

    \addplot [mark=pentagon, cyan] table [x expr=\coordindex, y={DIAR}]{\data};
    \addlegendentry{DIAR}

    \addplot [mark=square, orange] table [x expr=\coordindex, y={DQN+GNN}]{\data};
    \addlegendentry{DQN+GNN}

    \addplot [mark=star, black] table [x expr=\coordindex, y={IACR}]{\data};
    \addlegendentry{IACR}
    
    \addplot [mark=diamond, green!40!gray] table [x expr=\coordindex, y={OSPF}]{\data};
    \addlegendentry{OSPF}
    
    \addplot [mark=triangle, violet] table [x expr=\coordindex, y={RandomBL}]{\data};
    \addlegendentry{RB}
    
\end{axis}
\end{tikzpicture}

%% file: figures/NSFNET/Delay.tex
\pgfplotstableread[col sep=comma]{figures/NSFNET/nsfnet_delay_all_comps.csv}\data

\begin{tikzpicture}
\begin{axis}[
    width=\textwidth,
    height=10cm,
    xtick=data,
    xticklabels from table={\data}{N},
    legend pos=south east,
    legend cell align={left},
    ylabel={Delay [timesteps]},
    xlabel={Number of flows $N$},
    grid=both,
    grid style=dashed,
    ]
    
    \addplot [mark=triangle, violet] table [x expr=\coordindex, y={RandomBL}]{\data};
    \addlegendentry{RB}

    \addplot [mark=square, orange] table [x expr=\coordindex, y={DQN+GNN}]{\data};
    \addlegendentry{DQN+GNN}

    \addplot [mark=pentagon, cyan] table [x expr=\coordindex, y={DIAR}]{\data};
    \addlegendentry{DIAR}

    \addplot [mark=diamond, green!40!gray] table [x expr=\coordindex, y={OSPF}]{\data};
    \addlegendentry{OSPF}

    \addplot [mark=star, black] table [x expr=\coordindex, y={IACR}]{\data};
    \addlegendentry{IACR}

    \addplot [mark=o, blue] table [x expr=\coordindex, y={DIAMOND}]{\data};
    \addlegendentry{DIAMOND (ours)}
    
\end{axis}
\end{tikzpicture}

%% file: figures/Geant2/Rates.tex
\pgfplotstableread[col sep=comma]{figures/Geant2/geant2_rates_all_comps.csv}\data

\begin{tikzpicture}
\begin{axis}[
    width=\textwidth,
    height=10cm,
    xtick=data,
    xticklabels from table={\data}{N},
    legend pos=north east, 
    legend cell align={left},
    ylabel={Avg. Flows Rate [Mbps]},
    xlabel={Number of flows $N$},
    grid=both,
    grid style=dashed,
    ]
    
    \addplot [mark=o, blue] table [x expr=\coordindex, y={DIAMOND}]{\data};
    \addlegendentry{DIAMOND (ours)}

    \addplot [mark=pentagon, cyan] table [x expr=\coordindex, y={DIAR}]{\data};
    \addlegendentry{DIAR}

    \addplot [mark=square, orange] table [x expr=\coordindex, y={DQN+GNN}]{\data};
    \addlegendentry{DQN+GNN}

    \addplot [mark=star, black] table [x expr=\coordindex, y={IACR}]{\data};
    \addlegendentry{IACR}
    
    \addplot [mark=diamond, green!40!gray] table [x expr=\coordindex, y={OSPF}]{\data};
    \addlegendentry{OSPF}
    
    \addplot [mark=triangle, violet] table [x expr=\coordindex, y={RandomBL}]{\data};
    \addlegendentry{RB}
    
\end{axis}
\end{tikzpicture}

%% file: figures/Geant2/Delay.tex
\pgfplotstableread[col sep=comma]{figures/Geant2/geant2_delay_all_comps.csv}\data

\begin{tikzpicture}
\begin{axis}[
    width=\textwidth,
    height=10cm,
    xtick=data,
    xticklabels from table={\data}{N},
    legend pos=north west,  
    legend cell align={left},
    ylabel={Delay [timesteps]},
    xlabel={Number of flows $N$},
    grid=both,
    grid style=dashed,
    ]
    
    \addplot [mark=triangle, violet] table [x expr=\coordindex, y={RandomBL}]{\data};
    \addlegendentry{RB}

    \addplot [mark=diamond, green!40!gray] table [x expr=\coordindex, y={OSPF}]{\data};
    \addlegendentry{OSPF}

    \addplot [mark=square, orange] table [x expr=\coordindex, y={DQN+GNN}]{\data};
    \addlegendentry{DQN+GNN}

    \addplot [mark=pentagon, cyan] table [x expr=\coordindex, y={DIAR}]{\data};
    \addlegendentry{DIAR}

    \addplot [mark=star, black] table [x expr=\coordindex, y={IACR}]{\data};
    \addlegendentry{IACR}

    \addplot [mark=o, blue] table [x expr=\coordindex, y={DIAMOND}]{\data};
    \addlegendentry{DIAMOND (ours)}
    
\end{axis}
\end{tikzpicture}

%% file: figures/V30E50/Rates_equal.tex
\pgfplotstableread[col sep=comma]{figures/V30E50/random_equal_rates_V30E50.csv}\data
\begin{tikzpicture}
\begin{axis}[
    width=\textwidth,
    height=10cm,
    xtick=data,
    xticklabels from table={\data}{N},
    legend pos= south west, 
    legend cell align={left},
    ylabel={Avg. Flows Rate [Mbps]},
    xlabel={Number of flows $N$},
    grid=both,
    grid style=dashed,
    ymode=log,
    ]
    
    \addplot [mark=o, blue] table [x expr=\coordindex, y={DIAMOND}]{\data};
    \addlegendentry{DIAMOND (ours)}

    \addplot [mark=square, orange] table [x expr=\coordindex, y={DQN+GNN}]{\data};
    \addlegendentry{DQN+GNN}

    \addplot [mark=pentagon, cyan] table [x expr=\coordindex, y={DIAR}]{\data};
    \addlegendentry{DIAR}
    
    \addplot [mark=diamond, green!40!gray] table [x expr=\coordindex, y={OSPF}]{\data};
    \addlegendentry{OSPF}
    
    \addplot [mark=triangle, violet] table [x expr=\coordindex, y={RandomBL}]{\data};
    \addlegendentry{RB}

    \addplot [mark=star, black] table [x expr=\coordindex, y={IACR}]{\data};
    \addlegendentry{IACR}
    
\end{axis}
\end{tikzpicture}

%% file: figures/V30E50/Delay_equal.tex
\pgfplotstableread[col sep=comma]{figures/V30E50/random_equal_delay_V30E50.csv}\data

\begin{tikzpicture}
\begin{axis}[
    width=\textwidth,
    height=10cm,
    xtick=data,
    xticklabels from table={\data}{N},
    legend pos=north west,
    legend cell align={left},
    ylabel={Delay [timesteps]},
    xlabel={Number of flows $N$},
    grid=both,
    grid style=dashed,
    ymode=log,
    ]
    
    \addplot [mark=diamond, green!40!gray] table [x expr=\coordindex, y={OSPF}]{\data};
    \addlegendentry{OSPF}

    \addplot [mark=square, orange] table [x expr=\coordindex, y={DQN+GNN}]{\data};
    \addlegendentry{DQN+GNN}

    \addplot [mark=triangle, violet] table [x expr=\coordindex, y={RandomBL}]{\data};
    \addlegendentry{RB}

    \addplot [mark=pentagon, cyan] table [x expr=\coordindex, y={DIAR}]{\data};
    \addlegendentry{DIAR}

    \addplot [mark=star, black] table [x expr=\coordindex, y={IACR}]{\data};
    \addlegendentry{IACR}

    \addplot [mark=o, blue] table [x expr=\coordindex, y={DIAMOND}]{\data};
    \addlegendentry{DIAMOND (ours)}
    
\end{axis}
\end{tikzpicture}

%% file: figures/V30E50/Rates_steps.tex
\pgfplotstableread[col sep=comma]{figures/V30E50/random_steps_rates_V30E50.csv}\data
\begin{tikzpicture}
\begin{axis}[
    width=\textwidth,
    height=10cm,
    xtick=data,
    xticklabels from table={\data}{N},
    legend pos= south west, 
    legend cell align={left},
    ylabel={Avg. Flows Rate [Mbps]},
    xlabel={Number of flows $N$},
    grid=both,
    grid style=dashed,
    ymode=log,
    ]
    
    \addplot [mark=o, blue] table [x expr=\coordindex, y={DIAMOND}]{\data};
    \addlegendentry{DIAMOND (ours)}

    \addplot [mark=square, orange] table [x expr=\coordindex, y={DQN+GNN}]{\data};
    \addlegendentry{DQN+GNN}

    \addplot [mark=pentagon, cyan] table [x expr=\coordindex, y={DIAR}]{\data};
    \addlegendentry{DIAR}

    \addplot [mark=star, black] table [x expr=\coordindex, y={IACR}]{\data};
    \addlegendentry{IACR}
    
    \addplot [mark=diamond, green!40!gray] table [x expr=\coordindex, y={OSPF}]{\data};
    \addlegendentry{OSPF}
    
    \addplot [mark=triangle, violet] table [x expr=\coordindex, y={RandomBL}]{\data};
    \addlegendentry{RB}
    
\end{axis}
\end{tikzpicture}

%% file: figures/V30E50/Delay_steps.tex
\pgfplotstableread[col sep=comma]{figures/V30E50/random_steps_delay_V30E50.csv}\data

\begin{tikzpicture}
\begin{axis}[
    width=\textwidth,
    height=10cm,
    xtick=data,
    xticklabels from table={\data}{N},
    legend pos=north west,
    legend cell align={left},
    ylabel={Delay [timesteps]},
    xlabel={Number of flows $N$},
    grid=both,
    grid style=dashed,
    ymode=log,
    ]
    
    \addplot [mark=diamond, green!40!gray] table [x expr=\coordindex, y={OSPF}]{\data};
    \addlegendentry{OSPF}

    \addplot [mark=square, orange] table [x expr=\coordindex, y={DQN+GNN}]{\data};
    \addlegendentry{DQN+GNN}

    \addplot [mark=triangle, violet] table [x expr=\coordindex, y={RandomBL}]{\data};
    \addlegendentry{RB}

    \addplot [mark=pentagon, cyan] table [x expr=\coordindex, y={DIAR}]{\data};
    \addlegendentry{DIAR}

    \addplot [mark=star, black] table [x expr=\coordindex, y={IACR}]{\data};
    \addlegendentry{IACR}

    \addplot [mark=o, blue] table [x expr=\coordindex, y={DIAMOND}]{\data};
    \addlegendentry{DIAMOND (ours)}
    
\end{axis}
\end{tikzpicture}

%% file: figures/V30E50/Rates_rayleigh.tex
\pgfplotstableread[col sep=comma]{figures/V30E50/random_rayleigh_rates_V30_E50.csv}\data
\begin{tikzpicture}
\begin{axis}[
    width=\textwidth,
    height=10cm,
    xtick=data,
    xticklabels from table={\data}{N},
    legend pos= south west, 
    legend cell align={left},
    ylabel={Avg. Flows Rate [Mbps]},
    xlabel={Number of flows $N$},
    grid=both,
    grid style=dashed,
    ymode=log,
    ]
    
    \addplot [mark=o, blue] table [x expr=\coordindex, y={DIAMOND}]{\data};
    \addlegendentry{DIAMOND (ours)}

    \addplot [mark=square, orange] table [x expr=\coordindex, y={DQN+GNN}]{\data};
    \addlegendentry{DQN+GNN}

    \addplot [mark=pentagon, cyan] table [x expr=\coordindex, y={DIAR}]{\data};
    \addlegendentry{DIAR}

    \addplot [mark=star, black] table [x expr=\coordindex, y={IACR}]{\data};
    \addlegendentry{IACR}
    
    \addplot [mark=diamond, green!40!gray] table [x expr=\coordindex, y={OSPF}]{\data};
    \addlegendentry{OSPF}
    
    \addplot [mark=triangle, violet] table [x expr=\coordindex, y={RandomBL}]{\data};
    \addlegendentry{RB}
    
\end{axis}
\end{tikzpicture}

%% file: figures/V30E50/Delay_rayleigh.tex
\pgfplotstableread[col sep=comma]{figures/V30E50/random_rayleigh_delay_V30_E50.csv}\data

\begin{tikzpicture}
\begin{axis}[
    width=\textwidth,
    height=10cm,
    xtick=data,
    xticklabels from table={\data}{N},
    legend pos=north west,
    legend cell align={left},
    ylabel={Delay [timesteps]},
    xlabel={Number of flows $N$},
    grid=both,
    grid style=dashed,
    ymode=log,
    ]
    
    \addplot [mark=diamond, green!40!gray] table [x expr=\coordindex, y={OSPF}]{\data};
    \addlegendentry{OSPF}

    \addplot [mark=square, orange] table [x expr=\coordindex, y={DQN+GNN}]{\data};
    \addlegendentry{DQN+GNN}

    \addplot [mark=triangle, violet] table [x expr=\coordindex, y={RandomBL}]{\data};
    \addlegendentry{RB}

    \addplot [mark=pentagon, cyan] table [x expr=\coordindex, y={DIAR}]{\data};
    \addlegendentry{DIAR}

    \addplot [mark=star, black] table [x expr=\coordindex, y={IACR}]{\data};
    \addlegendentry{IACR}

    \addplot [mark=o, blue] table [x expr=\coordindex, y={DIAMOND}]{\data};
    \addlegendentry{DIAMOND (ours)}
    
\end{axis}
\end{tikzpicture}

%% file: figures/NB3R/NB3RRates.tex
\pgfplotstableread[col sep=comma]{figures/NB3R/NB3R_V50_E80_N20.csv}\data

\begin{tikzpicture}
\begin{axis}[
    width=\textwidth,
    height=10cm,
    legend pos=south east,
    legend cell align={left},
    ylabel={Average Flow Rates [Mbps]},
    xlabel={Time-step $t$},
    grid=both,
    grid style=dashed,
    ]

    \addplot [mark=o, blue, mark repeat=10,mark phase=0] table [x expr=\coordindex, y={rates}]{\data};
    \addlegendentry{Avg. Flows Rates}
    
\end{axis}
\label{fig:NB3R_convergance_rates}
\end{tikzpicture}

%% file: figures/NB3R/NB3RDelay.tex
\pgfplotstableread[col sep=comma]{figures/NB3R/NB3R_V50_E80_N20.csv}\data

\begin{tikzpicture}
\begin{axis}[
    width=\textwidth,
    height=10cm,
    legend pos=north east,
    legend cell align={left},
    ylabel={Max Delay [Hops]},
    xlabel={Time-step $t$},
    grid=both,
    grid style=dashed,
    ]
    
    \addplot [mark=triangle, violet, mark repeat=10,mark phase=0] table [x expr=\coordindex, y={delay}]{\data};
    \addlegendentry{Delay [Hops]}
        
\end{axis}
\label{fig:NB3R_convergance_delay}
\end{tikzpicture}

%% file: main.bbl
\begin{thebibliography}{10}
\providecommand{\url}[1]{#1}
\csname url@samestyle\endcsname
\providecommand{\newblock}{\relax}
\providecommand{\bibinfo}[2]{#2}
\providecommand{\BIBentrySTDinterwordspacing}{\spaceskip=0pt\relax}
\providecommand{\BIBentryALTinterwordstretchfactor}{4}
\providecommand{\BIBentryALTinterwordspacing}{\spaceskip=\fontdimen2\font plus
\BIBentryALTinterwordstretchfactor\fontdimen3\font minus
  \fontdimen4\font\relax}
\providecommand{\BIBforeignlanguage}[2]{{%
\expandafter\ifx\csname l@#1\endcsname\relax
\typeout{** WARNING: IEEEtran.bst: No hyphenation pattern has been}%
\typeout{** loaded for the language `#1'. Using the pattern for}%
\typeout{** the default language instead.}%
\else
\language=\csname l@#1\endcsname
\fi
#2}}
\providecommand{\BIBdecl}{\relax}
\BIBdecl

\bibitem{srikant2013communication}
R.~Srikant and L.~Ying, \emph{Communication networks: an optimization, control,
  and stochastic networks perspective}.\hskip 1em plus 0.5em minus 0.4em\relax
  Cambridge University Press, 2013.

\bibitem{gong2016distributed}
H.~Gong, L.~Fu, X.~Fu, L.~Zhao, K.~Wang, and X.~Wang, ``Distributed multicast
  tree construction in wireless sensor networks,'' \emph{IEEE Transactions on
  Information Theory}, vol.~63, no.~1, pp. 280--296, 2016.

\bibitem{ying2010combining}
L.~Ying, S.~Shakkottai, A.~Reddy, and S.~Liu, ``On combining shortest-path and
  back-pressure routing over multihop wireless networks,'' \emph{IEEE/ACM
  Transactions on Networking}, vol.~19, no.~3, pp. 841--854, 2010.

\bibitem{joo2011performance}
C.~Joo, ``On the performance of back-pressure scheduling schemes with
  logarithmic weight,'' \emph{IEEE transactions on wireless communications},
  vol.~10, no.~11, pp. 3632--3637, 2011.

\bibitem{amar2021online}
O.~Amar and K.~Cohen, ``Online learning for shortest path and backpressure
  routing in wireless networks,'' in \emph{IEEE International Symposium on
  Information Theory (ISIT)}, 2021, pp. 2702--2707.

\bibitem{liu2022routing}
Y.~Liu, H.~Mao, L.~Zhu, Z.~Xiao, Z.~Han, and X.-G. Xia, ``Routing and resource
  scheduling for air-ground integrated mesh networks,'' \emph{IEEE Transactions
  on Wireless Communications}, 2022.

\bibitem{tekin2011online}
C.~Tekin and M.~Liu, ``Online learning in opportunistic spectrum access: A
  restless bandit approach,'' in \emph{IEEE International Conference on
  Computer Communications (INFOCOM)}, 2011, pp. 2462--2470.

\bibitem{tekin2012online}
------, ``Online learning of rested and restless bandits,'' \emph{IEEE
  Transactions on Information Theory}, vol.~58, no.~8, pp. 5588--5611, 2012.

\bibitem{liu2012learning}
H.~Liu, K.~Liu, and Q.~Zhao, ``Learning in a changing world: Restless
  multiarmed bandit with unknown dynamics,'' \emph{IEEE Transactions on
  Information Theory}, vol.~59, no.~3, pp. 1902--1916, 2012.

\bibitem{cohen2014restless}
K.~Cohen, Q.~Zhao, and A.~Scaglione, ``Restless multi-armed bandits under
  time-varying activation constraints for dynamic spectrum access,'' in
  \emph{48th Asilomar Conference on Signals, Systems and Computers}.\hskip 1em
  plus 0.5em minus 0.4em\relax IEEE, 2014, pp. 1575--1578.

\bibitem{gafni2018learning}
T.~Gafni and K.~Cohen, ``Learning in restless multi-armed bandits using
  adaptive arm sequencing rules,'' in \emph{IEEE International Symposium on
  Information Theory (ISIT)}, 2018, pp. 1206--1210.

\bibitem{bistritz2018distributed}
I.~Bistritz and A.~Leshem, ``Distributed multi-player bandits-a game of thrones
  approach,'' in \emph{Advances in Neural Information Processing Systems},
  2018, pp. 7222--7232.

\bibitem{turgay2019exploiting}
E.~Tur{\u{g}}ay, C.~Bulucu, and C.~Tekin, ``Exploiting relevance for online
  decision-making in high-dimensions,'' \emph{IEEE Transactions on Signal
  Processing}, vol.~69, pp. 1438--1451, 2020.

\bibitem{yemini2020restless}
M.~Yemini, A.~Leshem, and A.~Somekh-Baruch, ``Restless hidden {Markov} bandit
  with linear rewards,'' in \emph{IEEE Conference on Decision and Control
  (CDC)}, 2020, pp. 1183--1189.

\bibitem{gafni2020learning}
T.~Gafni and K.~Cohen, ``Learning in restless multiarmed bandits via adaptive
  arm sequencing rules,'' \emph{IEEE Transactions on Automatic Control},
  vol.~66, no.~10, pp. 5029--5036, 2020.

\bibitem{gafni2022distributed}
------, ``Distributed learning over markovian fading channels for stable
  spectrum access,'' \emph{IEEE Access}, vol.~10, pp. 46\,652--46\,669, 2022.

\bibitem{gafni2022learning}
T.~Gafni, M.~Yemini, and K.~Cohen, ``Learning in restless bandits under
  exogenous global markov process,'' \emph{IEEE Transactions on Signal
  Processing}, vol.~70, pp. 5679--5693, 2022.

\bibitem{agrawal1995sample}
R.~Agrawal, ``Sample mean based index policies with {$O(log n)$} regret for the
  multi-armed bandit problem,'' \emph{Advances in Applied Probability}, pp.
  1054--1078, 1995.

\bibitem{auer2002finite}
P.~Auer, N.~Cesa-Bianchi, and P.~Fischer, ``Finite-time analysis of the
  multiarmed bandit problem,'' \emph{Machine learning}, vol.~47, no. 2-3, pp.
  235--256, 2002.

\bibitem{tabei2021multi}
G.~Tabei, Y.~Ito, T.~Kimura, and K.~Hirata, ``Multi-armed bandit-based routing
  method for in-network caching,'' in \emph{IEEE Asia-Pacific Signal and
  Information Processing Association Conference (APSIPA)}, 2021, pp.
  1899--1902.

\bibitem{wang2018deep}
S.~Wang, H.~Liu, P.~H. Gomes, and B.~Krishnamachari, ``Deep reinforcement
  learning for dynamic multichannel access in wireless networks,'' \emph{IEEE
  Transactions on Cognitive Communications and Networking}, vol.~4, no.~2, pp.
  257--265, 2018.

\bibitem{yu2019deep}
Y.~Yu, T.~Wang, and S.~C. Liew, ``Deep-reinforcement learning multiple access
  for heterogeneous wireless networks,'' \emph{IEEE Journal on Selected Areas
  in Communications}, vol.~37, no.~6, pp. 1277--1290, 2019.

\bibitem{naparstek2018deep}
O.~Naparstek and K.~Cohen, ``Deep multi-user reinforcement learning for
  distributed dynamic spectrum access,'' \emph{IEEE Transactions on Wireless
  Communications}, vol.~18, no.~1, pp. 310--323, 2019.

\bibitem{bokobza2023deep}
Y.~Bokobza, R.~Dabora, and K.~Cohen, ``Deep reinforcement learning for
  simultaneous sensing and channel access in cognitive networks,'' \emph{IEEE
  Transactions on Wireless Communications}, 2023.

\bibitem{liu2012adaptive}
K.~Liu and Q.~Zhao, ``Adaptive shortest-path routing under unknown and
  stochastically varying link states,'' in \emph{10th International Symposium
  on Modeling and Optimization in Mobile, Ad Hoc and Wireless Networks
  (WiOpt)}.\hskip 1em plus 0.5em minus 0.4em\relax IEEE, 2012, pp. 232--237.

\bibitem{tehrani2013distributed}
P.~Tehrani and Q.~Zhao, ``Distributed online learning of the shortest path
  under unknown random edge weights,'' in \emph{IEEE International Conference
  on Acoustics, Speech and Signal Processing (ICASSP)}, 2013, pp. 3138--3142.

\bibitem{he2013endhost}
T.~He, D.~Goeckel, R.~Raghavendra, and D.~Towsley, ``Endhost-based shortest
  path routing in dynamic networks: An online learning approach,'' in
  \emph{Proceedings of the IEEE INFOCOM}, 2013, pp. 2202--2210.

\bibitem{talebi2017stochastic}
M.~S. Talebi, Z.~Zou, R.~Combes, A.~Proutiere, and M.~Johansson, ``Stochastic
  online shortest path routing: The value of feedback,'' \emph{IEEE
  Transactions on Automatic Control}, vol.~63, no.~4, pp. 915--930, 2017.

\bibitem{xia2019reinforcement}
W.~Xia, C.~Di, H.~Guo, and S.~Li, ``Reinforcement learning based stochastic
  shortest path finding in wireless sensor networks,'' \emph{IEEE Access},
  vol.~7, pp. 157\,807--157\,817, 2019.

\bibitem{DRL+GNN}
P.~Almasan, J.~Su{\'{a}}rez{-}Varela, A.~Badia{-}Sampera, K.~Rusek,
  P.~Barlet{-}Ros, and A.~Cabellos{-}Aparicio, ``Deep reinforcement learning
  meets graph neural networks: An optical network routing use case,''
  \emph{CoRR}, vol. abs/1910.07421, 2019.

\bibitem{huang2021tsor}
Z.~Huang, Y.~Xu, and J.~Pan, ``Tsor: Thompson sampling-based opportunistic
  routing,'' \emph{IEEE Transactions on Wireless Communications}, vol.~20,
  no.~11, pp. 7272--7285, 2021.

\bibitem{Zhao2021}
L.~Zhao, W.~Zhao, A.~Hawbani, A.~Y. Al-Dubai, G.~Min, A.~Y. Zomaya, and
  C.~Gong, ``Novel online sequential learning-based adaptive routing for edge
  software-defined vehicular networks,'' \emph{IEEE Transactions on Wireless
  Communications}, vol.~20, no.~5, pp. 2991--3004, 2021.

\bibitem{amar2022online}
O.~Amar, I.~Sarfati, and K.~Cohen, ``An online learning approach to shortest
  path and backpressure routing in wireless networks,'' \emph{IEEE Access},
  2023.

\bibitem{Chai2020DIAR}
Y.~Chai and X.-J. Zeng, ``Delay- and interference-aware routing for wireless
  mesh network,'' \emph{IEEE Systems Journal}, vol.~14, no.~3, pp. 4119--4130,
  2020.

\bibitem{He2019JSRC}
S.~He, K.~Xie, K.~Xie, C.~Xu, and J.~Wang, ``Interference-aware multisource
  transmission in multiradio and multichannel wireless network,'' \emph{IEEE
  Systems Journal}, vol.~13, no.~3, pp. 2507--2518, 2019.

\bibitem{Waqas2022IACR}
A.~Waqas, H.~Mahmood, and N.~Saeed, ``Interference aware cooperative routing
  for edge computing-enabled 5g networks,'' \emph{IEEE Sensors Journal},
  vol.~22, no.~4, pp. 3777--3784, 2022.

\bibitem{liang2016deep}
S.~Liang and R.~Srikant, ``Why deep neural networks for function
  approximation?'' \emph{arXiv preprint arXiv:1610.04161}, 2016.

\bibitem{jiang2020dgn}
J.~Jiang, C.~Dun, T.~Huang, and Z.~Lu, ``Graph convolutional reinforcement
  learning,'' in \emph{ICLR}, 2020.

\bibitem{kortvelesy2022qgnn}
R.~Kortvelesy and A.~Prorok, ``Qgnn: Value function factorisation with graph
  neural networks,'' \emph{arXiv preprint arXiv:2205.13005}, 2022.

\bibitem{liu2020pic}
I.-J. Liu, R.~A. Yeh, and A.~G. Schwing, ``Pic: permutation invariant critic
  for multi-agent deep reinforcement learning,'' in \emph{Conference on Robot
  Learning}.\hskip 1em plus 0.5em minus 0.4em\relax PMLR, 2020, pp. 590--602.

\bibitem{NP_hard_routing}
M.~di~Ianni, ``Efficient delay routing,'' \emph{Theor. Comput. Sci.}, vol. 196,
  no. 1–2, p. 131–151, apr 1998.

\bibitem{babaee2010cross}
R.~Babaee and N.~C. Beaulieu, ``Cross-layer design for multihop wireless
  relaying networks,'' \emph{IEEE Transactions on Wireless Communications},
  vol.~9, no.~11, pp. 3522--3531, 2010.

\bibitem{GRU}
K.~Cho, B.~Van~Merri{\"e}nboer, C.~Gulcehre, D.~Bahdanau, F.~Bougares,
  H.~Schwenk, and Y.~Bengio, ``Learning phrase representations using rnn
  encoder-decoder for statistical machine translation,'' \emph{arXiv preprint
  arXiv:1406.1078}, 2014.

\bibitem{Shen2021Scalable}
Y.~Shen, Y.~Shi, J.~Zhang, and K.~B. Letaief, ``Graph neural networks for
  scalable radio resource management: Architecture design and theoretical
  analysis,'' \emph{IEEE Journal on Selected Areas in Communications}, vol.~39,
  no.~1, pp. 101--115, 2021.

\bibitem{liu2022survey}
X.~Liu, M.~Yan, L.~Deng, G.~Li, X.~Ye, D.~Fan, S.~Pan, and Y.~Xie, ``Survey on
  graph neural network acceleration: An algorithmic perspective,'' \emph{arXiv
  preprint arXiv:2202.04822}, 2022.

\bibitem{besta2022parallel}
M.~Besta and T.~Hoefler, ``Parallel and distributed graph neural networks: An
  in-depth concurrency analysis,'' \emph{arXiv preprint arXiv:2205.09702},
  2022.

\bibitem{Luo2022reram}
Y.~Luo, P.~Behnam, K.~Thorat, Z.~Liu, H.~Peng, S.~Huang, S.~Zhou, O.~Khan,
  A.~Tumanov, C.~Ding, and T.~Geng, ``Codg-reram: An algorithm-hardware
  co-design to accelerate semi-structured gnns on reram,'' in \emph{2022 IEEE
  40th International Conference on Computer Design (ICCD)}, 2022, pp. 280--289.

\bibitem{Zhang2021fccm}
B.~Zhang, R.~Kannan, and V.~Prasanna, ``Boostgcn: A framework for optimizing
  gcn inference on fpga,'' in \emph{2021 IEEE 29th Annual International
  Symposium on Field-Programmable Custom Computing Machines (FCCM)}, 2021, pp.
  29--39.

\bibitem{Zheng2022bytegnn}
\BIBentryALTinterwordspacing
C.~Zheng, H.~Chen, Y.~Cheng, Z.~Song, Y.~Wu, C.~Li, J.~Cheng, H.~Yang, and
  S.~Zhang, ``Bytegnn: Efficient graph neural network training at large
  scale,'' \emph{Proc. VLDB Endow.}, vol.~15, no.~6, p. 1228–1242, feb 2022.
  [Online]. Available: \url{https://doi.org/10.14778/3514061.3514069}
\BIBentrySTDinterwordspacing

\bibitem{REINFORCE}
R.~J. Williams, ``Simple statistical gradient-following algorithms for
  connectionist reinforcement learning,'' \emph{Mach. Learn.}, vol.~8, no.
  3–4, p. 229–256, may 1992.

\bibitem{BartoSutton}
R.~S.~a. Sutton, \emph{Reinforcement learning : an introduction}, second
  edition.~ed.\hskip 1em plus 0.5em minus 0.4em\relax Cambridge, Massachusetts
  :: The MIT Press,, 2018.

\bibitem{cohen2017distributed}
K.~Cohen, A.~Nedi{\'c}, and R.~Srikant, ``Distributed learning algorithms for
  spectrum sharing in spatial random access wireless networks,'' \emph{IEEE
  Transactions on Automatic Control}, vol.~62, no.~6, pp. 2854--2869, 2017.

\bibitem{young2020individual}
H.~P. Young, ``Individual strategy and social structure,'' in \emph{Individual
  Strategy and Social Structure}.\hskip 1em plus 0.5em minus 0.4em\relax
  Princeton University Press, 2020.

\bibitem{hajek1988cooling}
B.~Hajek, ``Cooling schedules for optimal annealing,'' \emph{Mathematics of
  operations research}, vol.~13, no.~2, pp. 311--329, 1988.

\bibitem{potential_games}
D.~Monderer and L.~S. Shapley, ``Potential games,'' \emph{Games and Economic
  Behavior}, vol.~14, no.~1, pp. 124--143, 1996.

\bibitem{blume1993statistical}
L.~E. Blume, ``The statistical mechanics of strategic interaction,''
  \emph{Games and economic behavior}, vol.~5, no.~3, pp. 387--424, 1993.

\bibitem{OSPF}
D.~Sidhu, T.~Fu, S.~Abdallah, R.~Nair, and R.~Coltun, ``Open shortest path
  first (ospf) routing protocol simulation,'' in \emph{Conference Proceedings
  on Communications Architectures, Protocols and Applications}, ser. SIGCOMM
  '93.\hskip 1em plus 0.5em minus 0.4em\relax New York, NY, USA: Association
  for Computing Machinery, 1993, p. 53–62.

\bibitem{mnih2015human}
V.~Mnih, K.~Kavukcuoglu, D.~Silver, A.~A. Rusu, J.~Veness, M.~G. Bellemare,
  A.~Graves, M.~Riedmiller, A.~K. Fidjeland, G.~Ostrovski, S.~Petersen,
  C.~Beattie, A.~Sadik, I.~Antonoglou, H.~King, D.~Kumaran, D.~Wierstra,
  S.~Legg, and D.~Hassabis, ``Human-level control through deep reinforcement
  learning,'' \emph{nature}, vol. 518, no. 7540, pp. 529--533, 2015.

\bibitem{Ju2012fullduplex}
H.~Ju, S.~Lim, D.~Kim, H.~V. Poor, and D.~Hong, ``Full duplexity in
  beamforming-based multi-hop relay networks,'' \emph{IEEE Journal on Selected
  Areas in Communications}, vol.~30, no.~8, pp. 1554--1565, 2012.

\bibitem{Stamatiou2010delayopt}
K.~Stamatiou and M.~Haenggi, ``The delay-optimal number of hops in poisson
  multi-hop networks,'' in \emph{2010 IEEE International Symposium on
  Information Theory}, 2010, pp. 1733--1737.

\bibitem{Li2018clustering}
J.~Li, B.~N. Silva, M.~Diyan, Z.~Cao, and K.~Han, ``A clustering based routing
  algorithm in iot aware wireless mesh networks,'' \emph{Sustainable Cities and
  Society}, vol.~40, pp. 657--666, 2018.

\bibitem{Vondrous2016performance}
O.~Vondrouš, Z.~Kocur, T.~Hégr, and O.~Slavíček, ``Performance evaluation
  of iot mesh networking technology in ism frequency band,'' in \emph{2016 17th
  International Conference on Mechatronics - Mechatronika (ME)}, 2016, pp.
  1--8.

\bibitem{Bisnik2006delay}
N.~Bisnik and A.~Abouzeid, ``Delay and throughput in random access wireless
  mesh networks,'' in \emph{2006 IEEE International Conference on
  Communications}, vol.~1, 2006, pp. 403--408.

\bibitem{NSFNET}
X.~Hei, J.~Zhang, B.~Bensaou, and C.-C. Cheung, ``Wavelength converter
  placement in least-load-routing-based optical networks using genetic
  algorithms,'' \emph{Journal of Optical Networking}, vol.~3, pp. 363--378, 05
  2004.

\bibitem{GEANT2}
F.~Barreto, E.~Wille, and L.~Nacamura, ``Fast emergency paths schema to
  overcome transient link failures in ospf routing,'' \emph{International
  Journal of Computer Networks \& Communications}, vol.~4, 04 2012.

\bibitem{van2016ddqn}
H.~Van~Hasselt, A.~Guez, and D.~Silver, ``Deep reinforcement learning with
  double q-learning,'' in \emph{Proceedings of the AAAI conference on
  artificial intelligence}, vol.~30, no.~1, 2016.

\end{thebibliography}
